%% file: ArxivSubmission.tex
\documentclass[runningheads]{llncs}

\include{preamble}

\include{defs}

\usepackage{comment}

\usepackage[width=122mm,left=12mm,paperwidth=146mm,height=193mm,top=12mm,paperheight=217mm]{geometry}

\begin{document}
\pagestyle{headings}
\mainmatter
\def\ECCVSubNumber{2942}  %

\title{6D Camera Relocalization in Ambiguous Scenes via Continuous Multimodal Inference} %

\titlerunning{Continuous Multimodal 6D Camera Relocalization}
\author{Mai Bui\inst{1} \,\,\,
Tolga Birdal\inst{2} \,\,\,
Haowen Deng\inst{1,3} \,\,\,
Shadi Albarqouni\inst{1,4} \,\,\,
\\Leonidas Guibas\inst{2} \,\,\,
Slobodan Ilic\inst{1,3} \,\,\,
Nassir Navab\inst{1,5}}
\authorrunning{M. Bui et al.}
\institute{
\,\inst{1} Technical University of Munich \,\,\,\,\,\, \inst{2} Stanford University\\
\inst{3} Siemens AG \,\,\,\,\,\, 
\inst{4} ETH Zurich \,\,\,\,\,\,
\inst{5} Johns Hopkins University}

\maketitle

\begin{abstract}
We present a multimodal camera relocalization framework that captures ambiguities and uncertainties with continuous mixture models defined on the manifold of camera poses. In highly ambiguous environments, which can easily arise due to symmetries and repetitive structures in the scene, computing one plausible solution (what most state-of-the-art methods currently regress) may not be sufficient. Instead we predict multiple camera pose hypotheses as well as the respective uncertainty for each prediction. Towards this aim, we use Bingham distributions, to model the orientation of the camera pose, and a multivariate Gaussian to model the position, with an end-to-end deep neural network. By incorporating a Winner-Takes-All training scheme, we finally obtain a mixture model that is well suited for explaining ambiguities in the scene, yet does not suffer from mode collapse, a common problem with mixture density networks. We introduce a new dataset specifically designed to foster camera localization research in ambiguous environments and exhaustively evaluate our method on synthetic as well as real data on both ambiguous scenes and on non-ambiguous benchmark datasets. We release our code and dataset under \href{https://multimodal3dvision.github.io}{multimodal3dvision.github.io}.
\end{abstract}%
\input{intro.tex}
\input{relatedwork.tex}
\input{bingham.tex}
\input{bingham_networks.tex}
\input{experiments.tex}
\input{conclusion.tex}

\clearpage
\input{acknowledgements}
\bibliographystyle{splncs04}

\input{ArxivSubmission.bbl}
\input{appendix.tex}

\end{document}

%% file: preamble.tex
\pdfoutput=1
\usepackage[utf8]{inputenc}
\usepackage{graphicx}
\usepackage{comment}
\usepackage{color}
\usepackage{amsmath,mathtools}

\usepackage{amsthm}	
\usepackage{amssymb}
\usepackage{enumerate}
\usepackage{subcaption}
\usepackage{algorithm} 
\usepackage{algpseudocode}
\usepackage{booktabs}
\usepackage{array}
\usepackage{bm}
\usepackage{url}
\usepackage{bbm}
\usepackage{enumitem}
\usepackage{stackengine}
\usepackage{multirow}
\stackMath
\usepackage{lipsum}
\usepackage{cite}
\usepackage{multirow}
\usepackage{tabularx}
\usepackage{xspace}

\usepackage[pagebackref=true,breaklinks=true,letterpaper=true,colorlinks,bookmarks=false]{hyperref}
\usepackage{cleveref}

\newcommand{\PBS}[1]{\let\temp=\\#1\let\\=\temp}
\newcommand{\RBS}{\let\\=\tabularnewline}

%% file: defs.tex
\newcommand{\etal}{\emph{et al}.}
\newcommand{\ie}{\emph{i.e. } }

\newcommand{\B}{\mathcal{B}}

\newcommand{\Img}{\mathbf{X}}
\newcommand{\ImgSet}{\mathcal{X}}

\newcommand{\Eye}{\mathbf{I}}

\newcommand{\q}{\mathbf{q}}
\newcommand{\Q}{\mathbf{Q}}
\newcommand{\qc}{\mathbf{\bar{q}}}
\newcommand{\tb}{\mathbf{t}}

\newcommand{\Hamil}{{\mathbb{H}}}

\newcommand{\V}{\mathbf{V}}

\newcommand{\Lo}{{\cal L}}

\newcommand{\Sp}{\mathbb{S}}

\newcommand{\R}{\mathbb{R}}

\DeclareMathOperator*{\argmin}{arg\min}
\DeclareMathOperator*{\argmax}{arg\max}

\renewcommand{\bm}[1]{\mathbf{#1}}

\newcommand{\eg}{\textit{e}.\textit{g}.}

\crefname{algorithm}{\text{Alg.}}{\text{Alg.}}
\crefname{assumption}{\textbf{H}}{\textbf{H}}
\crefname{equation}{\text{Eq}}{\text{Eq}}
\crefname{definition}{\text{Dfn.}}{\text{Dfn.}}
\crefname{lemma}{\text{Lemma}}{\text{Lemma}}
\crefname{dfn}{\text{Dfn.}}{\text{Dfn.}}
\crefname{thm}{\text{Thm.}}{\text{Thm.}}
\crefname{tab}{\text{Tab.}}{\text{Tab.}}
\crefname{fig}{\text{Fig.}}{\text{Fig.}}
\crefname{table}{\text{Tab.}}{\text{Tab.}}
\crefname{figure}{\text{Fig.}}{\text{Fig.}}
\crefname{section}{\text{Sec.}}{\text{Sec.}}

\Crefname{assumption}{\textbf{H}\hspace{-3pt}}{\textbf{H}\hspace{-3pt}}
\crefname{assumption}{\textbf{H}}{\textbf{H}}

\newcommand{\insertimageStar}[5]{ %
\begin{figure*}[#5]
\centering
\includegraphics[width=#1\linewidth, clip=true]{figures/#2}
\caption{#3}
\label{#4}
\end{figure*}
}

%% file: intro.tex
\section{Introduction}%
\label{sec:intro}
Camera relocalization is the term for determining the 6-DoF rotation and translation parameters of a camera with respect to a known 3D world. It is now a key technology in enabling a multitude of applications such as augmented reality, autonomous driving, human computer interaction and robot guidance, thanks to its extensive integration in simultaneous localization and mapping (SLAM)~\cite{durrant2006simultaneous,salas2013slam++,cadena2016past}, structure from motion (SfM)~\cite{ullman1979interpretation,schonberger2016structure}, metrology~\cite{birdal2016online} and visual localization~\cite{shotton2013scene,piasco2018survey}. For decades, vision scholars have worked on finding the unique solution of this problem~\cite{horaud1989analytic,horaud1989analytic,zeisl2015camera,sattler2015hyperpoints,hinterstoisser2012model,zakharov2019dpod,qi2019deep}. However, this trend is now witnessing a fundamental challenge. A recent school of thought has begun to point out that for highly complex and ambiguous real environments, obtaining a single solution for the location and orientation of a camera is simply not sufficient. This has led to a paradigm shift towards estimating a range of solutions, in the form of full probability distributions~\cite{birdal2018bayesian,arun2018probabilistic,birdal2019probabilistic} or at least solutions that estimate the uncertainty in orientation estimates~\cite{kendall2016modelling,manhardt2019explaining}.
 \begin{figure}
	\centering
	\includegraphics[width=\textwidth]{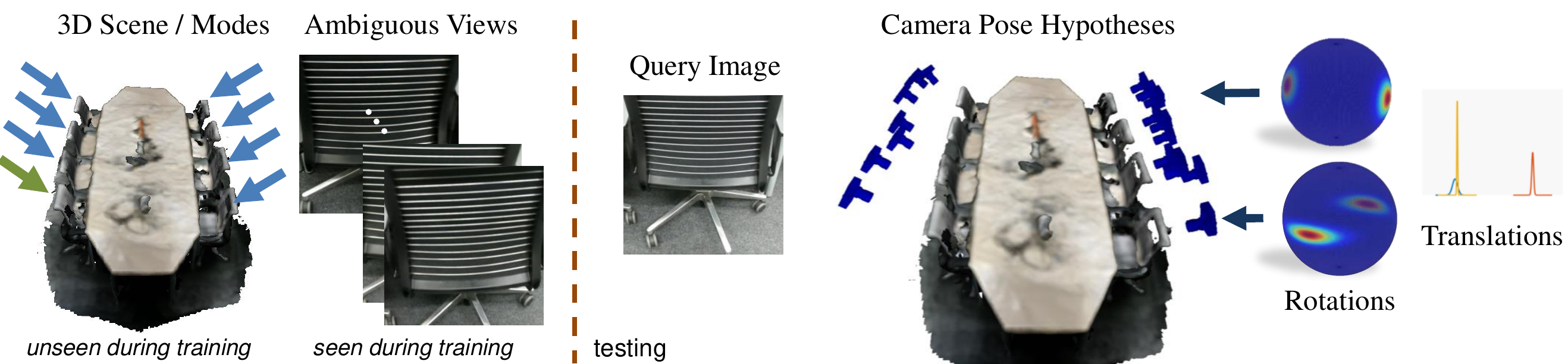}
	\caption{In a highly ambiguous environment, similar looking views can easily confuse current camera pose regression models and lead to incorrect localization results. Instead, given a query RGB image, our aim is to predict the possible modes as well as the associated uncertainties, which we model by the parameters of Bingham and Gaussian mixture models.}
	\label{fig:teaser}
\end{figure}
Thanks to advances in state-of-the-art machine learning, this important problem can now be tackled via data driven algorithms that are able to discover multi-modal and complex distributions, targeting the task at hand. 

In this paper, we devise a multi-hypotheses method, depicted in~\cref{fig:teaser}, for learning continuous mixture models on manifold valued rotations (parameterized by quaternions) and Euclidean translations that can explain uncertainty and ambiguity in 6DoF camera relocalization, while avoiding mode collapse~\cite{salimans2016improved}. In particular, we propose a direct regression framework utilizing a combination of antipodally symmetric Bingham~\cite{Bingham1974} and Gaussian probability distributions in order to deal with rotational and translational uncertainties respectively. Together, they are well suited to the geometric nature of $SE(3)$ pose representations. Using said distributions, we aim to build novel variational models that enable estimation of full covariances on the discrete modes to be predicted. For better exploration of the variational posterior, we extend the established particle based training approaches  ~\cite{makansi2019overcoming,rupprecht2017learning,makansi2019overcoming} to mixtures of Gaussians and Binghams. While these techniques only optimize the location of individual hypothesis to cover diverse modes, we additionally learn to predict associated variances on the manifold. We can then approximate the manifold valued posterior of our problem in a continuous fashion. Note that, to the best of our knowledge, such continuous distributions for multi-modal modeling of the network posteriors formed by the 6D pose parameters have not been explored previously. Our synthetic and real experiments demonstrate promising performance both under ambiguities and non-ambiguous scenes. Our method is also flexible in the sense that it can be used with a wide variety of backbone architectures.
In a nutshell, our contributions are:
\begin{enumerate}[noitemsep]
\item We provide a general framework for continuously modelling conditional density functions on quaternions using Bingham distributions, while explaining the translational uncertainty with multi-modal Gaussians.
\item We propose a new training scheme for deep neural networks that enables the inference of a diverse set of modes and related concentration parameters as well as the prior weights for the mixture components.
\item We exhaustively evaluate our method on existing datasets, demonstrating the validity of our approach. Additionally, we create a new highly ambiguous camera relocalization dataset, which we use to showcase the quality of results attained by our algorithm and provide a thorough study on camera localization in ambiguous scenes. Further, we make our dataset publicly available with the hope to foster future research.
\end{enumerate}

%% file: relatedwork.tex
\section{Prior Art}
\label{sec:related}
6D camera relocalization is a very well studied topic with a vast literature~\cite{shotton2013scene,kendall2015posenet,brahmbhatt2018geometry,brachmann2018learning,brachmann2017dsac,bui2018bmvc,feng20166d,massiceti2017random}. Our work considers the uncertainty aspect and this is what we focus on here: We first review the uncertainty estimation in deep networks, and subsequently move to uncertainty in 6D pose and relocalization.
\paragraph{\textbf{Characterizing the Posterior in Deep Networks}}
Typical CNNs~\cite{Simonyan14c,he2016deep} are over-confident in their predictions~\cite{guo2017calibration,zolfaghari2019learning}. Moreover, these networks tend to approximate the conditional averages of the target data~\cite{bishop1994mixture}. These undesired properties render the immediate outputs of those networks unsuitable for the quantification of calibrated uncertainty. This has fostered numerous works as we will summarize in the following. \emph{Mixture Density Networks (MDN)}~\cite{bishop1994mixture} is the pioneer to model the conditional distribution by predicting the parameters of a mixture model. Yet, it is repeatedly reported that optimizing for general mixture models suffers from mode collapse and numerical instabilities~\cite{cui2019multimodal,makansi2019overcoming}.
These issues can to a certain extent be alleviated by using Dropout~\cite{gal2016dropout} as a Bayesian approximation, but even for moderate dimensions these methods still face difficulties in capturing multiple modes. Instead, the more tailored \emph{Winner Takes All} (WTA)~\cite{guzman2012multiple,firman2018diversenet} as well as \emph{Relaxed-WTA} (RWTA)~\cite{rupprecht2017learning} try to capture the multimodal posterior in the $K$-best hypotheses predictions of the network. \emph{Evolving-WTA} (EWTA)~\cite{makansi2019overcoming} further avoids the inconsistencies related to the WTA losses.
Though, a majority of these works consider only low dimensional posterior with the assumption of a Euclidean space, whereas we consider a 7D non-Euclidean highly non-convex posterior.

\paragraph{\textbf{Uncertainty in 6D}}
Initial attempts to capture the uncertainty of camera relocalization involved Random Forests (RF)~\cite{breiman2001random}. Valentin~\etal~\cite{valentin2015exploiting} stored GMM components at the leaves of a scene coordinate regression forest~\cite{shotton2013scene}. The modes are obtained via a mean shift procedure, and the covariance is explained by a 3D Gaussian. A similar approach later considered the uncertainty in object coordinate labels~\cite{brachmann2016uncertainty}. It is a shortcoming of RF that both of these approaches require hand crafted depth features. Moreover, their uncertainty is on the correspondences and not on the final camera pose. Thus a costly RANSAC~\cite{fischler1981random} is required to propagate the uncertainty in the leaves to the camera pose.

Only recently, Manhardt~\etal~\cite{manhardt2019explaining} localized a camera against a known object under rotational ambiguities arising due to symmetries or self-occlusions. They extended the RTWA~\cite{rupprecht2017learning} to deal with the 3D rotations using quaternions. This method can only yield discrete hypotheses not continuous density estimates. Similarly, the pose estimation network of Pitteri~\etal~\cite{pitteri2019object} explicitly considered axis-symmetric objects whose pose cannot be uniquely determined. Likewise, Corona~\etal~\cite{corona2018pose} addressed general rotational symmetries. All of these works require extensive knowledge about the object and cannot be extended to the scenario of localizing against a scene without having a 3D model. Note that the latter two works cannot handle the case of self-symmetry and~\cite{corona2018pose} additionally requires a dataset of symmetry-labeled objects, an assumption unlikely to be fulfilled in real applications.

Bayesian PoseNet~\cite{kendall2016modelling} was one of the first works to model uncertainty for the 6D relocalization problem. It leveraged Dropout~\cite{gal2016dropout} to sample the posterior as a way to enable approximate probabilistic pose inference. Although in theory this method can generate discrete samples from the multi-modal distribution, in practice, as we will demonstrate, the Monte Carlo scheme tends to draw samples around a single mode. This method also suffers from the large errors associated to PoseNet~\cite{kendall2015posenet} itself. 
The successive VidLoc~\cite{clark2017vidloc} adapted MDNs~\cite{bishop1994mixture} to model and predict uncertainty for the 6D relocalization problem. Besides the reported issues of MDNs, VidLoc incorrectly modeled the rotation parameters using Gaussians and lacked the demonstrations of uncertainty on rotations. Contrarily, in this work we devise a principled method using Bingham distributions~\cite{Bingham1974} that are well suited to the double covering nature of unit quaternions.
HydraNet~\cite{peretroukhin2019probabilistic} provided calibrated uncertainties on the $SO(3)$ group, but assumed a unimodal posterior that is centered on the naive $\R^4$-mean of predicted quaternions.

Our work is inspired by~\cite{prokudin2018deep}, where a variational auto-encoder~\cite{kingma2013auto} is learnt to approximate the posterior of $SO(2)$ modeled by von Mises mixtures~\cite{mardia2009directional}. Though, it is not trivial to tweak and generalize~\cite{prokudin2018deep} to the continuous, highly multi-modal and multi-dimensional setting of 6D camera relocaliztion. This is what we precisely contribute in this work. Note that we are particularly interested in the \textit{aleatoric} uncertainty (noise in the observations) and leave the \textit{epistemic} (noise in the model) part as a future work~\cite{kendall2017uncertainties}. 

%% file: bingham.tex
\section{The Bingham Distribution}
Derived from a zero-mean Gaussian, the Bingham distribution~\cite{Bingham1974} (BD) is an antipodally symmetric probability distribution conditioned to lie on $\Sp^{d-1}$ with probability density function (PDF) $\B : \Sp^{d-1} \rightarrow R$:
\begin{equation}
\label{eq:bingham}
\B(\mathbf{x}; \mathbf{\Lambda}, \mathbf{V}) = (1/F) \exp(\mathbf{x}^T\mathbf{V}\mathbf{\Lambda}\mathbf{V}^T\mathbf{x}) 
= (1/F) \exp\big(\sum\nolimits_{i=1}^d \lambda_i(\mathbf{v}_i^T \mathbf{x})^2 \big)
\end{equation}
where $\mathbf{V} \in \mathbb{R}^{d\times d}$ is an orthogonal matrix $(\mathbf{V}\mathbf{V}^T = \mathbf{V}^T\mathbf{V} = \mathbf{I}_{d\times d})$ describing the orientation, $\mathbf{\Lambda} \in R^{d\times d}$ is called the \textit{concentration matrix} with $0 \geq \lambda_1 \geq \cdots \geq \lambda_{d-1}$: $\mathbf{\Lambda}=\text{diag}(\begin{bmatrix} 0 & \lambda_1 & \lambda_2 & \dots & \lambda_{d-1} \end{bmatrix}).$

It is easy to show that adding a multiple of the identity matrix $\mathbf{I}_{d\times d}$ to $\mathbf{V}$ does not change the distribution \cite{Bingham1974}. Thus, we conveniently force the first entry of $\mathbf{\Lambda}$ to be zero. Moreover, since it is possible to swap columns of $\mathbf{\Lambda}$, we can build $\mathbf{V}$ in a sorted fashion. This allows us to obtain \textit{the mode} very easily by taking the first column of $\mathbf{V}$. Due to its antipodally symmetric nature, the mean of the distribution is always zero.
$F$ in~\cref{eq:bingham} denotes the \textit{the normalization constant} dependent only on $\mathbf{\Lambda}$ and is of the form:
\begin{equation}
F \triangleq |S_{d-1}| \cdot {}_{1}F_1 \Big({1}/{2},\,{d}/{2},\, \mathbf{\Lambda}\Big),
\end{equation}
where $|S_{d-1}|$ is the surface area of the $d$-sphere and ${}_{1}F_1$ is a confluent hypergeometric function of matrix argument~\cite{carl1995,kurz2013recursive}. The computation of $F$ is not trivial. In practice, following Glover~\cite{glover2014quaternion}, this quantity as well as its gradients are approximated by tri-linear interpolation using a pre-computed look-up table over a predefined set of possible values in $\bm{\Lambda}$, lending itself to differentiation~\cite{kume2005saddlepoint,kurz2017directional}. 

\paragraph{\textbf{Relationship to quaternions}}
The antipodal symmetry of the PDF makes it amenable to explain the topology of quaternions, i. e., $\B(\mathbf{x}; \cdot) = \B(-\mathbf{x}; \cdot)$ holds for all $\mathbf{x} \in \Sp^{d-1}$. 
In 4D when $\lambda_1=\lambda_2=\lambda_3$, one can write $\mathbf{\Lambda}=\text{diag}([1,0,0,0])$. In this case, Bingham density relates to the dot product of two quaternions $\q_1 \in \Hamil_1\triangleq\mathbf{x}$ and the mode of the distribution, say $\qc_2\in \Hamil_1$. This induces a metric of the form: $d_{\text{bingham}}=d(\q_1,\qc_2)=(\q_1 \cdot \qc_2)^2 = \text{cos}(\theta/2)^2$.

Bingham distributions have been extensively used to represent distributions on unit quaternions ($\Hamil_1$)~\cite{glover2012monte,kurz2013recursive,glover2014,birdal2018bayesian,birdal2020measure}; however, to the best of our knowledge, never for the problem we consider here. 
\paragraph{\textbf{Constructing a Bingham distribution on a given mode}}
Creating a Bingham distribution on any given mode $\q \in \Hamil_1$ requires finding a set of vectors orthonormal to $\q$. This is a frame bundle $\Hamil_1 \rightarrow \mathcal{F}\Hamil_1$ composed of four unit vectors: the mode and its orthonormals. We follow Birdal~\etal~\cite{birdal2018bayesian} and use the \textit{parallelizability} of unit quaternions to define the orthonormal basis $\mathbf{V}: \Hamil_1 \mapsto \R^{4 \times 4} $:
\begin{align}
\label{eq:V}
\mathbf{V}(\q) \triangleq 
\begin{bmatrix}
q_1 & -q_2 				& -q_3 				&  \phantom{-}q_4 \\
q_2 & \phantom{-}q_1 	& \phantom{-}q_4 	&  \phantom{-}q_3\\
q_3 & -q_4 				& \phantom{-}q_1 	&  -q_2\\
q_4 & \phantom{-}q_3 	& -q_2				&   -q_1
\end{bmatrix}.
\end{align}

It is easy to verify that the matrix valued function $\mathbf{V}(\q)$ is orthonormal for every $\q \in \Hamil_1$. 
$\mathbf{V}(\q)$ further gives a convenient way to represent quaternions as matrices paving the way to linear operations, such as quaternion multiplication or orthonormalization without the Gram-Schmidt.
\paragraph{\textbf{Relationship to other representations}}
Note that geometric~\cite{barfoot2014associating} or measure theoretic~\cite{falorsi2019reparameterizing}, there are multitudes of ways of defining probability distributions on the Lie group of 6D rigid transformations~\cite{haarbach2018survey}. A choice would be to define Gaussian distribution on the Rodrigues vector (or exponential coordinates)~\cite{murray1994} where the geodesics are straight lines~\cite{morawiec1996rodrigues} or the use of Concentrated Gaussian distributions~\cite{bourmaud2015continuous} on matrices of SE(3). However, as our purpose is not tracking but direct regression, in this work we favor quaternions as continuous and minimally redundant parameterizations without singularities~\cite{grassia1998,busam2017camera} and use the Bingham distribution that is well suited to their topology. We handle the redundancy $\mathbf{q}\equiv-\mathbf{q}$ by mapping all the rotations to the northern hemisphere. 

%% file: bingham_networks.tex
\section{Proposed Model}
\label{sec:model}
We now describe our model for uncertainty prediction following~\cite{prokudin2018deep}. We consider the situation where we observe an input image $\Img\in \R^{W\times H \times 3}$ and assume the availability of a predictor function $\mu_{\bm{\Gamma}}(\Img) : \R^{W\times H \times 3} \mapsto \Hamil_1$ parameterized by $\bm{\Gamma}=\{\bm{\Gamma}_i\}$. Note that predicting entities that are non-Euclidean easily generalizes to prediction of Euclidean quantities such as translations e.g. $\tb\in\R^3$. For the sake of conciseness and clarity, we will omit the translations and concentrate on the rotations. Translations modeled via Gaussians will be precised later on.
\paragraph{\textbf{The unimodal case}} We momentarily assume that $\mu_{\bm{\Gamma}}(\cdot)$, or short $\mu(\cdot)$, can yield the correct values of the absolute camera rotations $\q_i\in\Hamil_1$ with respect to a common origin, admitting a non-ambiguous prediction, hence a posterior of single mode. We use the predicted rotation to set the most likely value (mode) of a BD:
\begin{align}
 p_{\bm{\Gamma}} (\q \,|\, \Img ; \bm{\Lambda}) = (1/F)~\text{exp } \big(\q^\top \V_{\mu} \bm{\Lambda} \V_{\mu}^\top \q \big),
\end{align}
and let $\q_i$ differ from this value up to the extent determined by $\bm{\Lambda}=\{\lambda_i\}$. For the sake of brevity we use $\V_\mu\equiv\V(\mu(\Img))$, the orthonormal basis aligned with the predicted quaternion $\mu(\Img)$ and as defined in~\cref{eq:V}.

\insertimageStar{1}{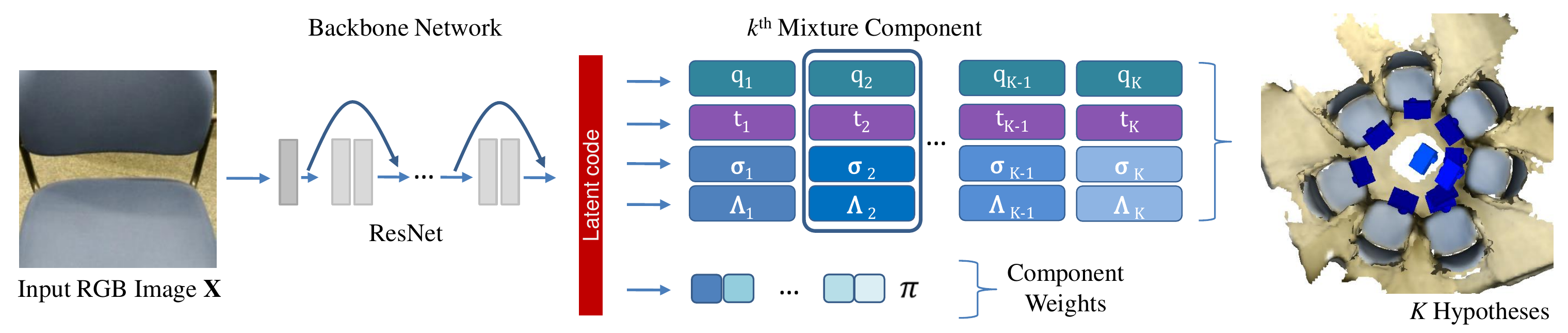}{Forward pass of our network. For an input RGB image we predict \textit{K} camera pose hypotheses as well as Bingham concentration parameters, Gaussian variances and component weights to obtain a mixture model.}{fig:arch}{t!}

While for certain applications, fixing $\bm{\Lambda}$ can work, in order to capture the variation in the input, it is recommended to adapt $\bm{\Lambda}$~\cite{prokudin2018deep}. Thus, we introduce it among the unknowns. To this end we define the function ${\bm{\Lambda}}_{\bm{\Gamma}}(\Img)$ or in short $\bm{\Lambda}_{\bm{\Gamma}}$ for computing the concentration values depending on the current image and the parameters ${\bm{\Gamma}}$. Our final model for the unimodal case reads:
\begin{align}
\label{eq:posterior}
 p_{\bm{\Gamma}} (\q \,|\, \Img) &= \frac{\text{exp } \big(\q^\top \V(\mu(\Img)) {\bm{\Lambda}}_{\bm{\Gamma}}(\Img) \V(\mu(\Img))^\top \q \big)}{F({\bm{\Lambda_{\bm{\Gamma}}}(\Img)})}
= \frac{\text{exp } \big(\q^\top \V_{\mu} \bm{\Lambda}_{\bm{\Gamma}} \V_{\mu}^\top \q \big)}{F(\bm{\Lambda_{\bm{\Gamma}}})}
\end{align}
The latter follows from the short-hand notations and is included for clarity.
Given a collection of observations i.e., images $\ImgSet=\{\Img_i\}$ and associated rotations $\Q=\{\q_i\}$, where $i=1,...,N$, the parameters of $\mu_{\bm{\Gamma}}(\Img)$ and ${\bm{\Lambda}}_{\bm{\Gamma}}(\Img)$ can be obtained simply by maximizing the log-likelihood:
\begin{align}
\bm{\Gamma}^\star &= \argmax_{\bm{\Gamma}} \text{ log }\Lo_u (\bm{\Gamma} | \ImgSet, \Q)\\
\text{log }\Lo_u (\bm{\Gamma} | \ImgSet, \Q) &= \sum\limits_{i=1}^N \q_i^\top \V_{\bm{\mu}} \bm{\Lambda}_{\bm{\Gamma}} \V_{\bm{\mu}}^\top \q_i- \sum\limits_{i=1}^N {\text{log} F\big(\bm{\Lambda}_{\bm{\Gamma}}\big)}.
\end{align}
Note once again that $\bm{\Lambda}_{\bm{\Gamma}}\equiv \bm{\Lambda}_{\bm{\Gamma}}(\Img_i)$ and $\V_\mu\equiv\V(\mu(\Img_i))$. If $\bm{\Lambda}_{\bm{\Gamma}}$ were to be fixed as in~\cite{prokudin2018deep}, the term on the right would have no effect and minimizing that loss would correspond to optimizing the Bingham log-likelihood. To ensure $0 \geq \lambda_1 \geq \cdots \geq \lambda_{d-1}$, we parameterize $\boldsymbol{\lambda}$ by $\lambda_1$ and the positive offsets $e_2,..., e_{d-1}$ such that 
$\lambda_{k} = \lambda_{k-1} - e_{k}$ where $k=2,...,d-1$.
This allows us to make an ordered prediction from the network.

\paragraph{\textbf{Extension to finite Bingham Mixture Models (BMM)}}
Ambiguities present in the data requires us to take into account the multimodal nature of the posterior. To achieve this, we now extend the aforementioned model to Bingham Mixture Models~\cite{riedel2016multi}. For the finite case, we use $K$ different components associated with $K$ mixture weights $\pi_j(\Img, \bm{\Gamma})$ for $j = 1, ..., K$. With each component being a Bingham distribution, we can describe the density function as
\begin{align}
	\label{eq:mixture}
	P_{\bm{\Gamma}} (\q \,|\, \Img) = \sum_{j=1}^{K} \pi_j(\Img, \bm{\Gamma}) p_{\bm{\Gamma}j} (\q \,|\, \Img),
\end{align}
where $p_{\bm{\Gamma}j} (\q \,|\, \Img)$ are the $K$ component distributions and $\pi_j(\Img, \bm{\Gamma})$ the mixture weights s.t.  $\sum_{j} \pi_j(\Img, \bm{\Gamma}) = 1$.
The model can again be trained by maximizing the log-likelihood, but this time of the mixture model~\cite{ley2018directional,yamaji2016genetic}
\begin{align}
\bm{\Gamma}^\star &= \argmax_{\bm{\Gamma}} \text{ log }\Lo_m (\bm{\Gamma} | \ImgSet, \Q)\\
\text{log }\Lo_m (\bm{\Gamma} | \ImgSet, \Q) &= \sum\limits_{i=1}^N \text{log} \sum_{j=1}^{K} \pi_j(\Img_i, \bm{\Gamma}) p_{\bm{\Gamma}j} (\q_i \,|\, \Img_i).
\end{align}

\section{Deeply modeling $\mu(\cdot)$ and $\Lambda(\cdot)$}
Following up on the recent advances, we jointly model $\mu(\cdot)$ and $\Lambda(\cdot)$ by a deep residual network~\cite{he2016deep}. $\bm{\Gamma}$ denotes the entirety of the trainable parameters. On the output we have \textbf{fourteen} quantities per density: four for the mode quaternion, three for translation, three for $\bm{\Lambda}$ the Bingham concentration, three for variances of the multivariate Gaussian and one for the weight $\pi_j(\cdot)$. In total our $K$ mixture components result in $K\times 14$ output entities. Our architecture is shown in~\cref{fig:arch} and we provide further details in the suppl. document. While a typical way to train our network is through simultaneously regressing the output variables, this is known to severely harm the accuracy~\cite{rupprecht2017learning}. Instead we exploit modern approaches to training in presence of ambiguities as we detail in what follows.

\paragraph{\textbf{MHP training scheme}}
Due to the increased dimensionality, in practice training our variational network in an unconstrained manner is likely to suffer from mode collapse, where all the heads concentrate around the same prediction. To avoid this and obtain a diverse set of modes, instead of training all branches equally by maximizing the log-likelihood of the mixture model, we follow the multi-hypotheses schemes of~\cite{rupprecht2017learning,makansi2019overcoming} and train our model using a WTA loss function, for each branch maximizing the log-likelihood of a unimodal distribution,
\begin{equation}
	\bm{\Gamma}^\star = \argmax_{\bm{\Gamma}}  \sum_{i=1}^{N}\sum_{j=1}^{K} w_{ij} \log \Lo_u (\bm{\Gamma} | \Img_i, \q_i)), 
\end{equation}
according to the associated weights $w_{ij}$ for each of the $k$ hypotheses. In this work, we compute the weights $w_{ij}$ during training following RWTA~\cite{rupprecht2017learning} as
\begin{equation}
	w_{ij} = \begin{cases}
	1-\epsilon,& \text{if } j = \argmin_k | \q_{i} - \hat{\q}_{ik} | \\
	\frac{\epsilon}{K-1},              & \text{otherwise}
	\end{cases}, 
\end{equation}
where $\hat{\q}_{ik}$ is the predicted mode of a single Bingham distribution.
Note that WTA~\cite{guzman2012multiple} would amount to updating only the branch of the best hypothesis and EWTA~\cite{makansi2019overcoming} the top $k$ branches closest to the ground truth. However, for our problem, we found RWTA to be a more reliable machinery. Finally, to obtain the desired continuous distribution, we train the weights of our Bingham mixture model using the following loss function:
\begin{equation}
	\Lo_{\pi}(\bm{\Gamma} | \ImgSet, \Q) = \sum_{i=1}^{N} \sum_{j=1}^{K} \sigma(\hat{\pi}_j(\Img_i, \bm{\Gamma}), y_{ij}),
\end{equation}
where $\sigma$ is the cross-entropy, $\hat{\pi}(\Img, \bm{\Gamma})$ the predicted weight of the neural network and $y_{ij}$ the associated label of the mixture model component given as
\begin{equation}
y_{ij} = \begin{cases}
1,& \text{if } j = \argmin_{k} | \q_{i} - \hat{\q}_{ik} |\\
0,              & \text{otherwise}
\end{cases}.
\end{equation}
Our final loss, therefore, consists of the weighted likelihood for a unimodal distribution of each branch and the loss of our mixture weights, $\Lo_{\pi}(\bm{\Gamma} | \ImgSet, \Q)$.  

\paragraph{\textbf{Incorporating translations}} We model translations $\{\tb_i\in\R^{3}\}_i$ by the standard Gaussian distributions with covariances $\{\bm{\Sigma}_i\in\R^{3\times 3}\succeq 0\}_i$. Hence, we use the ordinary MDNs~\cite{bishop1994mixture} to handle them. Yet, once again, during training we apply the MHP scheme explained above to avoid mode collapse and diversify the predictions. In practice, we first train the network to predict the translation and its variance. Then, intuitively, recovering the associated rotation should be an easier task, after which we fine-tune the network on all components of the distribution. Such split has already been shown to be prosperous in prior work~\cite{Deng2019}.
\paragraph{\textbf{Inference}}

Rather than reporting the conditional average which can result in label blur, we propose to obtain a single best estimate according to the weighted mode, where we choose the best mixture component according to its mixture weight and pick the mode as a final prediction.

We finally measure the uncertainty of the prediction according to the entropy of the resulting Bingham and Gaussian distributions, given as
\begin{align}
H_B = \text{log} F - \bm{\Lambda} \frac{ \nabla F(\bm{\Lambda})}{F}, \quad\text{ and }\quad H_G = \frac{c}{2} + \frac{c}{2} \text{log}(2\pi) + \frac{1}{2} \text{log}(|\bm{\Sigma}|),
\end{align}
respectively, where $c=3$ the dimension of the mean vector of the Gaussian. For a given image we first normalize the entropy values over all pose hypotheses, and finally obtain a measure of (un)certainty as the sum of both rotational ($H_B$) and translational ($H_G$) normalized entropy.  
\paragraph{\textbf{Implementation details}}
We implement our method in Python using PyTorch library~\cite{pytorch}. Following the current state-of-the-art direct camera pose regression methods, we use a \textit{ResNet-34}~\cite{he2016deep} as our backbone network architecture, followed by fully-connected layers for rotation and translation, respectively. We follow a projected ADAM optimization~\cite{kingma2014adam}, where the predicted quaternions are normalized during training. 
We add additional fully-connected layers with \textit{softplus} activation function, to ensure positivity of the Bingham concentration parameters and Gaussian variances. To satisfy the convention, the Bingham concentration parameters are then negated. For mixture model predictions, we use $K=50$ pose hypotheses. We run the ADAM optimizer with an exponential learning rate decay and train each model for 300 epochs and a batch size of 20 images.
We provide further details of training in the supplementary material.

%% file: experiments.tex
\section{Experimental Evaluation}
\label{sec:exp}
\input{tables/sota.tex}
When evaluating our method we consider two cases: (1) camera relocalization in non-ambiguous scenes, where our aim is to not only predict the camera pose, but the posterior of both rotation and translation that can be used to associate each pose with a measure of uncertainty; (2) we create a highly ambiguous environment, where similar looking images are captured from very different viewpoints. We show the problems current regression methods suffer from in handling such scenarios and in contrast show the merit of our proposed method. 
\paragraph{\textbf{Error metrics}} Note that, under ambiguities a best mode is unlikely to exist. In those cases, as long as we can generate a hypothesis that is close to the Ground Truth (GT), our network is considered successful. For this reason, in addition to the standard metrics and the weighted mode, we will also speak of the so called \textit{Oracle} error, assuming an oracle that is able to choose the best of all predictions: the one closest to the GT. In addition, we report the \textit{Self-EMD} (SEMD)~\cite{makansi2019overcoming}, the earth movers distance~\cite{rubner2000earth} of turning a multi-modal distribution into a unimodal one. With this measure we can evaluate the diversity of predictions, where the unimodal distribution is chosen as the predicted mode of the corresponding method. Note that this measure by itself does not give any indication about the accuracy of the prediction.
\paragraph{\textbf{Datasets}} 
In addition to the standard datasets of 7-Scenes \cite{shotton2013scene} and Cambridge Landmarks \cite{kendall2015posenet}, we created synthetic as well as real datasets, that are specifically designed to contain repetitive structures and allow us to assess the real benefits of our approach. For synthetic data we render table models from 3DWarehouse\footnote{https://3dwarehouse.sketchup.com/} and create camera trajectories, e.g. a circular movement around the object, such that ambiguous views are ensured to be included in our dataset. Specifically we use a \textit{dining table} and a \textit{round table} model with discrete modes of ambiguities. In addition, we create highly ambiguous real scenes using Google Tango and the graph-based SLAM approach RTAB-Map \cite{labbe2019rtab}. We acquire RGB and depth images as well as distinct ground truth camera trajectories for training and testing. We also reconstruct those scenes. However, note that only the RGB images and corresponding camera poses are required to train our model and the reconstructions are used for visualization only. In particular, our training and test sets consist of 2414 and 1326 frames, respectively. Note that our network sees a single pose label per image. We provide further details, visualizations and evaluations in our supplementary material.
\paragraph{\textbf{Baselines and SoTA}} 
We compare our approach to current state-of-the-art direct camera pose regression methods, PoseNet \cite{kendall2017geometric} and MapNet \cite{brahmbhatt2018geometry}, that output a single pose prediction. More importantly, we assess our performance against two state-of-the-art approaches, namely BayesianPoseNet \cite{kendall2016modelling} and VidLoc \cite{clark2017vidloc}, that are most related to our work and predict a distribution over the pose space by using dropout and mixture density networks, respectively.
We further include the \textit{unimodal} predictions as well as BMMs trained using mixture density networks~\cite{bishop1994mixture,Gilitschenski2020Deep} as baselines. We coin the latter Bingham-MDN or in short \textit{BMDN}.

\begin{figure}[t]\footnotesize 
	\centering
	\begin{subfigure}[b]{0.49\textwidth}
		\includegraphics[width=\textwidth]{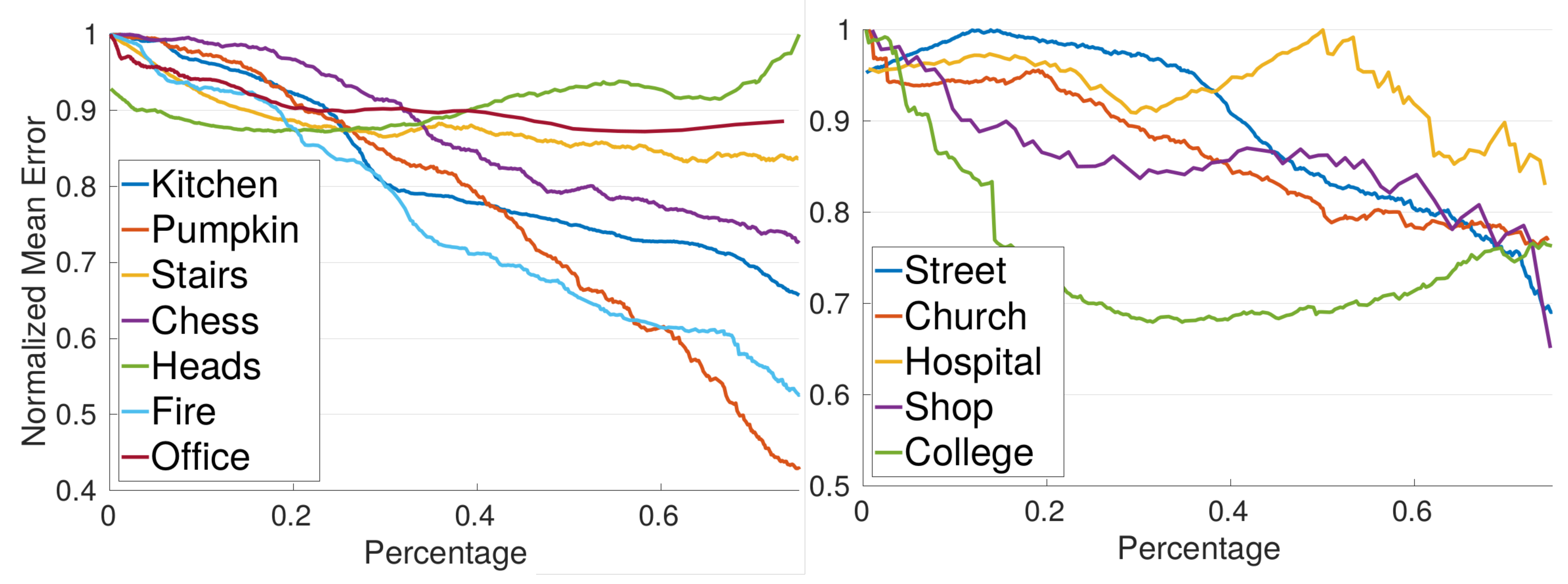}
		\caption{\footnotesize Rot. Uncertainty}
	\end{subfigure}\hfill
	\begin{subfigure}[b]{0.49\textwidth}
		\includegraphics[width=\textwidth]{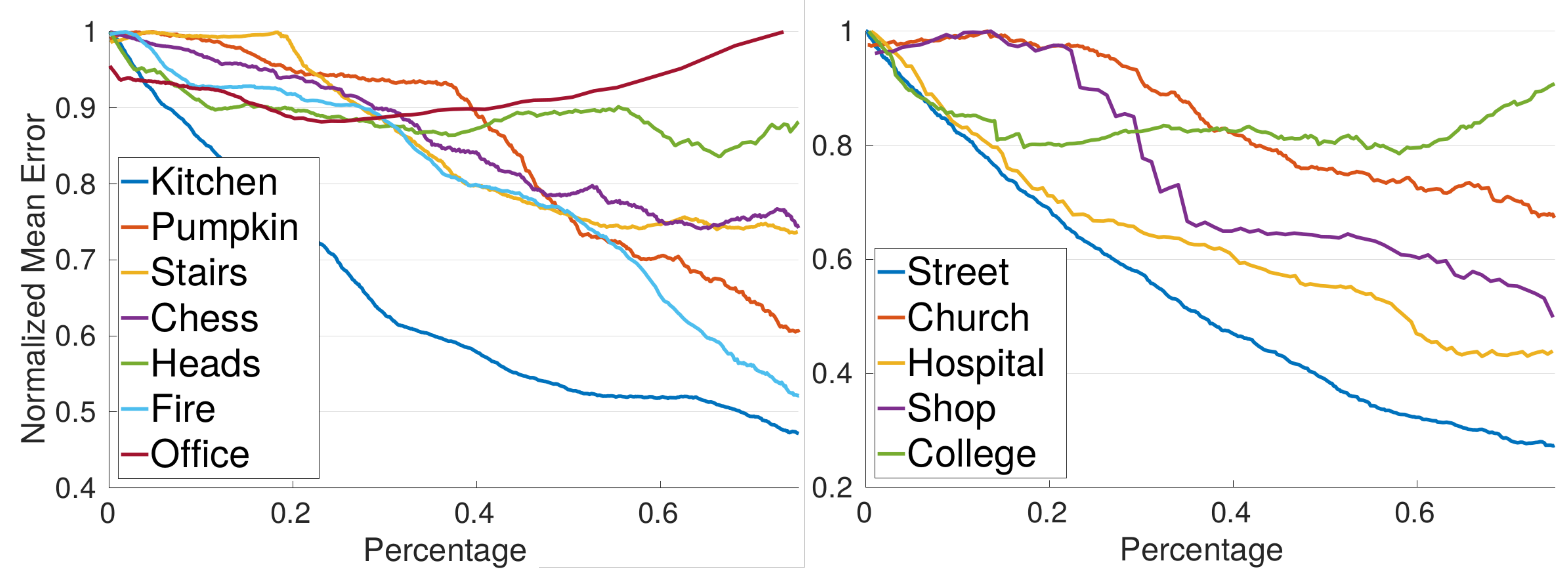}
		\caption{\footnotesize Trans. Uncertainty}
	\end{subfigure}
	\caption{Uncertainty evaluation on the 7-Scenes and Cambridge Landmarks datasets, showing the correlation between predicted uncertainty and pose error. Based on the entropy of our predicted distribution uncertain samples are gradually removed. We observe that as we remove the uncertain samples the overall error drops indicating a strong correlation between our predictions and the actual erroneous estimations.}
	\label{fig:uncertainty}
\end{figure}

\subsection{Evaluation in non-ambiguous scenes}
We first evaluate our method on the publicly available 7-Scenes \cite{shotton2013scene} and Cambridge Landmarks \cite{kendall2015posenet} datasets. As most of the scenes contained in these datasets do not show highly ambiguous environments, we consider them to be non-ambiguous. Though, we can not guarantee that some ambiguous views might arise in these datasets as well, such as in the \textit{Stairs} scene of the 7-Scenes dataset. Both datasets have extensively been used to evaluate camera pose estimation methods. Following the state-of-the-art, we report the median rotation and translation errors, the results of which can be found in~\cref{table:results}. In comparison to methods that output a single pose prediction (\eg~PoseNet \cite{kendall2017geometric} and MapNet \cite{brahmbhatt2018geometry}), our methods achieves similar results. Yet, our network provides an additional piece of information that is the uncertainty. On the other hand, especially in translation our method outperforms uncertainty methods, namely BayesianPoseNet \cite{kendall2016modelling} and VidLoc \cite{clark2017vidloc}, on most scenes.
\paragraph{\textbf{Uncertainty evaluation}}
One benefit of our method is that we can use the resulting variance of the predicted distribution as a measure of uncertainty in our predictions. The resulting correlation between pose error and uncertainty can be seen in~\cref{fig:uncertainty}, where we gradually remove the most uncertain predictions and plot the mean error for the remaining samples. The inverse correlation between the actual errors vs our confidence shows that whenever our algorithm labels a prediction as uncertain it is also likely to be a bad estimate. 
\input{tables/ambiguous_dataset.tex}

It has been shown that current direct camera pose regression methods still have difficulties in generalizing to views that differ significantly from the camera trajectories seen during training \cite{sattler2019understanding}. However, we chose to focus on another problem these methods have to face and analyze the performance of direct regression methods in a highly ambiguous environment. In this scenario even similar trajectories can confuse the network and easily lead to wrong predictions, for which our method proposes a solution.
\subsection{Evaluation in ambiguous scenes}
We start with quantitative evaluations on our synthetic as well as real scenes before showing qualitative results. We compare our method to PoseNet and BayesianPoseNet, which we refer to as MC-Dropout. In comparison, we replace the original network architecture by a ResNet, that has been shown to improve the performance of direct camera pose regression methods \cite{brahmbhatt2018geometry}.

\paragraph{\textbf{Quantitative evaluations}}
Due to the design of our synthetic table scenes, we know that there are two and four possible modes for each image in \textit{dining} and \textit{round} table scenes respectively. Hence, we analyze the predictions of our model by computing the accuracy of correctly detected modes of the true posterior. A mode is considered as found if there exists one pose hypothesis that falls into a certain rotational (5$^\circ$) and translational (10\% of the diameter of GT camera trajectory) threshold of it. In the dining-table, MC-Dropout obtains an accuracy of 50\%, finding one mode for each image, whereas the accuracy of Ours-RWTA on average achieves 96\%. On round-table, our model shows an average detection rate of $99.1\%$, in comparison to $24.8\%$ of MC-Dropout.

On our real scenes, we report the recall, where a pose is considered to be correct if both the rotation and translation errors are below a pre-defined threshold.~\cref{table:ambigious} shows the accuracy of our baseline methods in comparison to ours for various thresholds. Especially on our \textit{Meeting Table} scene, it can be seen that the performance of direct camera pose regression methods that suffer from mode collapse drops significantly due to the presence of ambiguities in the scene. Thanks to the diverse mode predictions of Ours-RWTA, which is indicated by the high Oracle accuracy as well as the high SEMD shown in~\cref{tab:semd}, we are able to improve upon our baseline predictions.
\begin{figure*}[t]
	\centering
	\includegraphics[width=\textwidth]{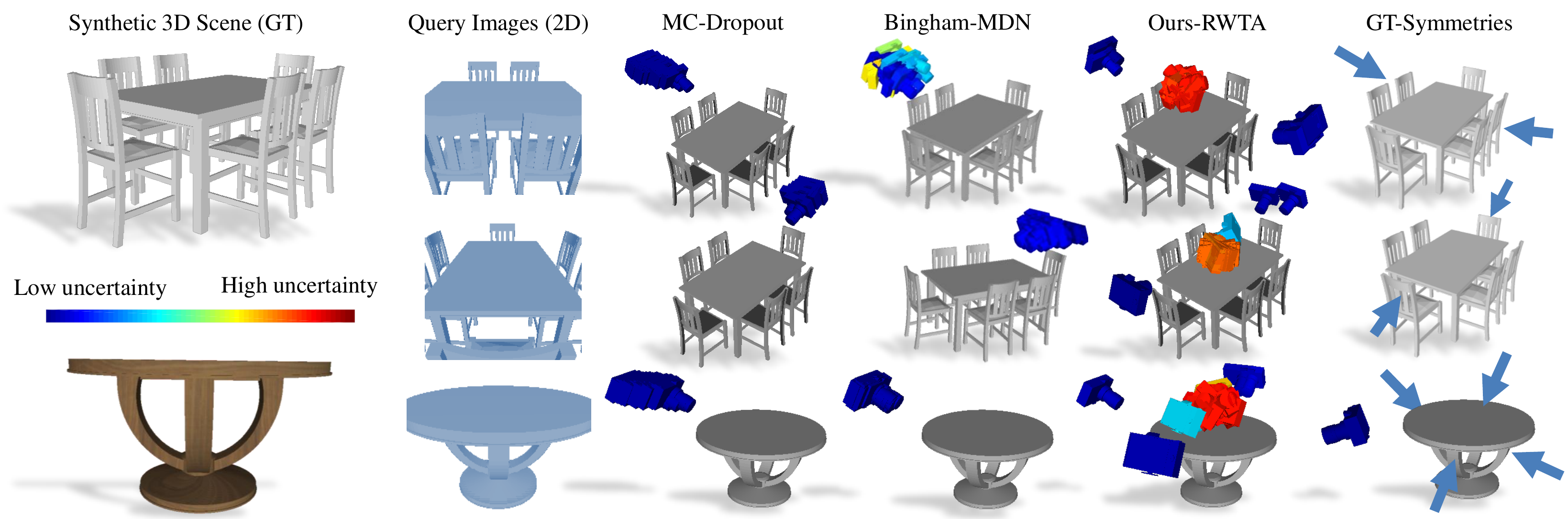}
	\caption{Qualitative results on our synthetic \textit{dining} and \textit{round table} datasets. Camera poses are colored by uncertainty.  
	Viewpoints are adjusted for best perception.}
	\label{fig:synth}
\end{figure*}\input{tables/SEMD}
Further, by a semi-automatic labeling procedure detailed in our suppl. material, we are able to extract GT modes for the \textit{Blue Chairs} and \textit{Meeting Table} scenes. This way, we can evaluate the entire set of predictions against the GT.~\cref{table:mode_detection} shows the percentage of correctly detected modes for our method in comparison to MC-Dropout when evaluating with these GT modes. The results support our qualitative observations, that MC-Dropout suffers from mode collapse such that even with increasing threshold, the number of detected modes does not increase significantly. 
\paragraph{\textbf{Qualitative evaluations}}
Qualitative results of our proposed model on our synthetic table datasets are shown in~\cref{fig:synth}. MC-Dropout as well as our finite mixture model, \textit{Bingham-MDN}, suffer from mode collapse. In comparison, the proposed MHP model is able to capture plausible, but diverse modes as well as associated uncertainties. In contrast to other methods that obtain an uncertainty value for one prediction, we obtain uncertainty values for each hypothesis. This way, we could easily remove non-meaningful predictions, that for example can arise in the WTA and RWTA training schemes. Resulting predicted Bingham distributions are visualized in~\cref{fig:bingham_plot}, where we marginalize over the angle component. %

\begin{figure*}[t]
	\centering
	\includegraphics[width=\textwidth]{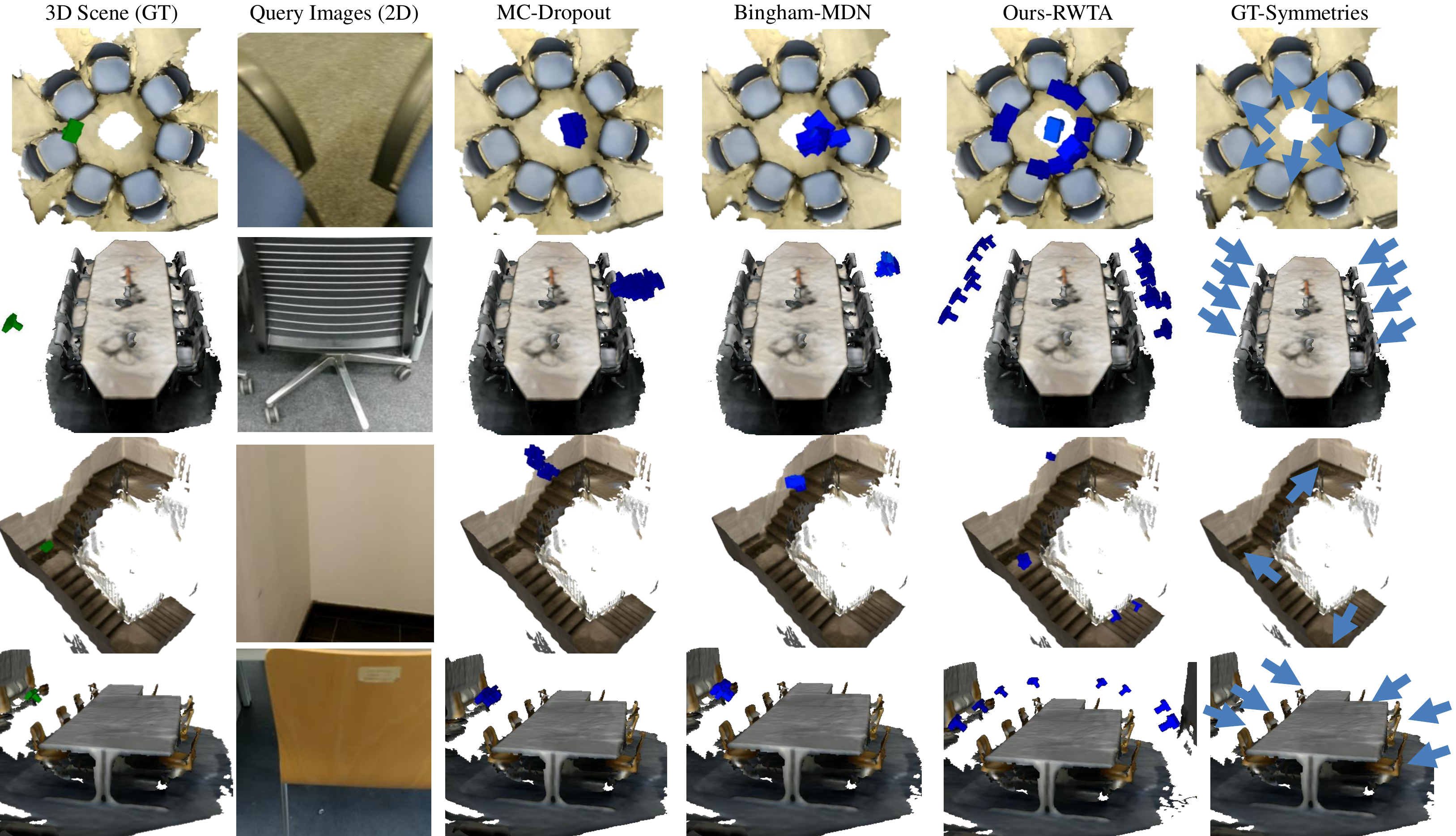}
	\caption{Qualitative results in our ambiguous dataset. For better visualization, camera poses have been pruned by their uncertainty values.}
	\label{fig:real}
\end{figure*}

\cref{fig:real} shows qualitative results on our ambiguous real scenes. Again, MC-Dropout and Bingham-MDN suffer from mode collapse. Moreover, these methods are unable to predict reasonable poses given highly ambiguous query images. This effect is most profound in our \textit{Meeting Table} scene, where due to its symmetric structure the predicted camera poses fall on the opposite side of the ground truth one. 
\input{tables/mhpANDmodes}
\subsection{Ablation Studies}
We now evaluate the effect of the two distinct parts of our network: multi-hypotheses predictions and the backbone architectures. We provide more ablations in our supplementary material.
\paragraph{\textbf{Multiple hypothesis estimation}}
Recently,~\cite{makansi2019overcoming} proposed EWTA, an evolving version of WTA, to alleviate the collapse problems of the RWTA training schemes proposed in~\cite{rupprecht2017learning}. Updating the top $k$ hypotheses instead of only the best one, EWTA increases the number of hypotheses that are actually used during training, resulting in fewer wrong mode predictions that do not match the actual distribution. We evaluated the different versions of MHP training schemes for our particular application for which the results can be found in~\cref{table:ewta}.
As it is not straightforward how $k$ should be chosen in EWTA, we 1) start with $k=K$, where $K$ is the number of hypotheses and gradually decrease $k$ until $k=1$ (as proposed in \cite{makansi2019overcoming}) and 2) start with the best half hypotheses, i.e. $k=0.5 \cdot K$. We set $K=50$ in our experiments. In our setting, we have found this parameter to strongly influence the accuracy of our model. Meanwhile, the wrong predictions are showing very high uncertainty so that, if desired, they can easily be removed. Therefore, we chose to remain with RWTA to train our models. This implicitly admits $k=1$. Note, however, that these conclusions were drawn from experimental results on our datasets, such that the optimal choice of training scheme remains application dependant.
\input{tables/backbone_average}
\paragraph{\textbf{Backbone network}} To evaluate the effect of different network architectures on our model, we change the backbone network of ours and the SoTA baseline methods. As most of the recent SoTA image based relocalization methods \cite{balntas2018relocnet, brahmbhatt2018geometry, peretroukhin2019probabilistic} use a version of ResNet, we compare between ResNet variants: ResNet-18, ResNet-34 and ResNet-50 and Inception-v3~\cite{szegedy2016rethinking}. All the networks are initialized from an ImageNet~\cite{deng2009imagenet} pre-trained model. We report our findings in~\cref{table:backbone}. Naturally all methods are slightly dependant on the features that serve as input to the final pose regression layers.
However, regardless of the backbone network used, Ours-RWTA shows, on average, superior performance over the baseline methods.

%% file: tables/sota.tex
\begin{table*}[t]
	\begin{center}
		\caption{Evaluation in non-ambiguous scenes, displayed is the median rotation and translation error. (Numbers for MapNet on the Cambridge Landmarks dataset are taken from \cite{sattler2019understanding}). BPN depicts Bayesian-PoseNet~\cite{brahmbhatt2018geometry}. \textit{Uni} and \textit{BMDN} refer to our unimodal version and Bingham-MDN respectively.}
		\vspace{2mm}
		\resizebox{\textwidth}{!}{
			\begin{tabular}{lccccccc|ccccc}
				Dataset & \multicolumn{7}{c}{7-Scenes} & \multicolumn{4}{c}{Cambridge Landmarks}\\
				\noalign{\smallskip}
				 $[^\circ~/~\text{m}]$& Chess & Fire & Heads & Office & Pumpkin & Kitchen & Stairs  & Kings & Hospital & Shop & St. Marys & Street\\
				\noalign{\smallskip}
				\midrule
				\noalign{\smallskip}
				PoseNet & $4.48$/$0.13$& $\textbf{11.3}$/$0.27$ & $13.0$/$0.17$ & $5.55$/$0.19$ & $4.75$/$0.26$ & $5.35$/$0.23$ & $12.4$/$0.35$ & $\textbf{1.04}$/$0.88$ & $\textbf{3.29}$/$3.2$ & $\textbf{3.78}$/$0.88$ & $\textbf{3.32}$/$1.57$ & $25.5$/$20.3$\\
				MapNet & $\textbf{3.25}$/$\textbf{0.08}$ & $11.69$/$0.27$ & $13.2$/$0.18$ & $\textbf{5.15}$/$\textbf{0.17}$ & $\textbf{4.02}$/$\textbf{0.22}$ & $\textbf{4.93}$/$0.23$ & $12.08$/$0.3$ & $1.89$/$1.07$ & $3.91$/$1.94$ & $4.22$/$1.49$ & $4.53$/$2.0$ &-\\
				\midrule
				\noalign{\smallskip}
				BPN & $7.24$/$0.37$ & $13.7$/$0.43$ & $\textbf{12.0}$/$0.31$ & $8.04$/$0.48$ & $7.08$/$0.61$ & $7.54$/$0.58$ & $13.1$/ $0.48$ & $4.06$/$1.74$ & $5.12$/$2.57$ & $7.54$/$1.25$ & $8.38$/$2.11$&-\\
				VidLoc& -/$0.18$ & -/$\textbf{0.26}$ & -/$0.14$ & -/$0.26$ & -/$0.36$ & -/$0.31$ & -/$\textbf{0.26}$ & - & - & - & - &-\\
				Uni & $4.97$/$0.1$ & $12.87$/$0.27$ & $14.05$/$\textbf{0.12}$ & $7.52$/$0.2$ & $7.11$/$0.23$ & $8.25$/$\textbf{0.19}$ & $13.1$/$0.28$ & $1.77$/$0.88$ & $3.71$/$\textbf{1.93}$ & $4.74$/$\textbf{0.8}$ & $6.19$/$1.84$ & $\textbf{24.1}$/$16.8$\\
				BMDN & $4.35$/$0.1$ & $11.86$/$0.28$ & $12.76$/$\textbf{0.12}$ & $6.55$/$0.19$ & $6.9$/$\textbf{0.22}$ & $8.08$/$0.21$ & $\textbf{9.98}$/$0.31$ & $2.08$/$\textbf{0.83}$ & $3.64$/$2.16$ & $4.93$/$0.92$ & $6.03$/$\textbf{1.37}$ & $36.9$/$\textbf{9.7}$\\
				\bottomrule
			\end{tabular}
		}
		\label{table:results}
	\end{center}
\end{table*}

%% file: tables/ambiguous_dataset.tex
\begin{table*}[t]
	\begin{center}
		\caption{ Ratio of correct poses on our ambiguous scenes for several thresholds.}
		\vspace{1mm}
		\resizebox{\textwidth}{!}{		
			\begin{tabular}{lcccccc|cc}
				& Threshold & \vtop{\hbox{\strut PoseNet}\hbox{\strut ~~~\cite{kendall2015posenet}}}& Uni. & BMDN & \vtop{\hbox{\strut MC-Dropout }\hbox{\strut ~~~~~~~\cite{kendall2016modelling}}}
			    & Ours-RWTA & \vtop{\hbox{\strut MC-Dropout }\hbox{\strut ~~~~Oracle}} & \vtop{\hbox{\strut Ours-RWTA }\hbox{\strut ~~~~Oracle}} \\
				\noalign{\smallskip}
				\midrule
				\noalign{\smallskip}
				& 10$^\circ$ / 0.1m & 0.19 & 0.29 & 0.24 & \textbf{0.39} & 0.35 & 0.40 & \textbf{0.58} \\
				Blue Chairs (A) &15$^\circ$ / 0.2m & 0.69 & 0.73 & 0.75 & 0.78 & \textbf{0.81} & 0.90 & \textbf{0.94}\\
				&20$^\circ$ / 0.3m & \textbf{0.90} & 0.86 & 0.80 & 0.88 & 0.82 & 0.95 & \textbf{1.00}\\
				\midrule
				&10$^\circ$ / 0.1m & 0.0 & 0.02 & 0.01 & 0.04 & \textbf{0.05} & \textbf{0.13} & 0.12 \\
				Meeting Table (B) &15$^\circ$ / 0.2m & 0.05 & 0.12 & 0.07 & 0.13 & \textbf{0.28} & 0.27 & \textbf{0.56}\\
				&20$^\circ$ / 0.3m & 0.10 & 0.19 & 0.10 & 0.22 & \textbf{0.39} & 0.32 & \textbf{0.78}\\
				\midrule
				&10$^\circ$ / 0.1m & 0.14 & 0.11 & 0.04 & 0.13 & \textbf{0.18} & \textbf{0.27} & 0.19 \\
				Staircase (C) &15$^\circ$ / 0.2m & 0.45 & 0.48 & 0.15 & 0.32 & \textbf{0.50} & \textbf{0.54}& 0.53\\
				&20$^\circ$ / 0.3m & 0.60 & 0.62 & 0.25 & 0.49 & \textbf{0.68} & 0.70 & \textbf{0.74}\\
				\midrule
				&10$^\circ$ / 0.1m & 0.07 & 0.06 & 0.06 & 0.02 & \textbf{0.09} & \textbf{0.16} & 0.09 \\
				Staircase Extended (D) &15$^\circ$ / 0.2m & 0.31 & 0.26 & 0.21 & 0.14 & \textbf{0.39} & \textbf{0.45} & 0.40\\
				&20$^\circ$ / 0.3m & 0.49 & 0.41 & 0.32 & 0.31 & \textbf{0.58} & \textbf{0.64} & \textbf{0.64}\\
				\midrule
				&10$^\circ$ / 0.1m & \textbf{0.37} & 0.11 & 0.06 & 0.18 & 0.35 & \textbf{0.46} &  0.36\\
				Seminar Room (E)&15$^\circ$ / 0.2m & 0.81 & 0.36 & 0.23 & 0.57 & \textbf{0.83} & \textbf{0.85} & 0.83\\
				&20$^\circ$ / 0.3m & 0.90 & 0.57 & 0.40 & 0.78 & \textbf{0.95} & 0.90 & \textbf{0.95}\\
				\midrule\midrule
				&10$^\circ$ / 0.1m & 0.15 & 0.12 & 0.08 & 0.15 & \textbf{0.20} & \textbf{0.28} &  0.27\\
				Average &15$^\circ$ / 0.2m & 0.46 & 0.39 & 0.28 & 0.39 & \textbf{0.56} & 0.60 & \textbf{0.65}\\
				&20$^\circ$ / 0.3m & 0.60 & 0.53 & 0.37 & 0.54 & \textbf{0.68} & 0.70 & \textbf{0.82}\\
				\bottomrule
			\end{tabular}
		}
		\label{table:ambigious}
	\end{center}
\end{table*}

%% file: tables/SEMD.tex
\addtocounter{figure}{-1}
\begin{figure}[ht]
\centering
\begin{minipage}[b]{0.47\textwidth}
\centering
\setlength{\tabcolsep}{0.8pt}
\begin{tabular}{lccccc}
			Method/Scene & A & B & C & D & E \\
			\midrule
			MC-Dropout & 0.06 & 0.11 & 0.13 & 0.26 & 0.10 \\
			Ours-RWTA & \textbf{1.19} & \textbf{2.13} & \textbf{2.04} & \textbf{3.81} & \textbf{1.70} \\
			\bottomrule
		\end{tabular}
		\captionof{table}{\footnotesize SEMD of our method and MC-Dropout indicating highly diverse predictions by our method in comparison to the baseline.}
		\label{tab:semd}
\end{minipage}\hfill
\begin{minipage}[b]{0.50\textwidth}
\centering
	\begin{subfigure}[b]{0.19\textwidth}
		\includegraphics[width=\textwidth]{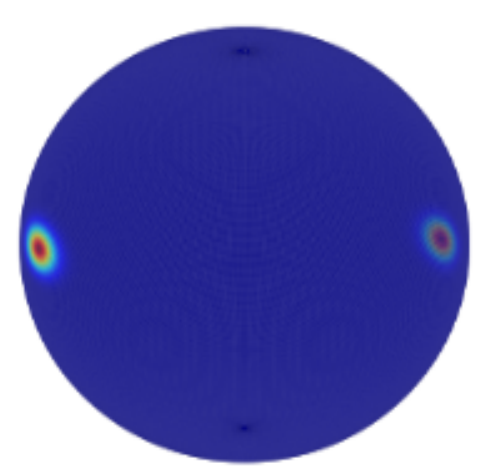}
		\caption{}
	\end{subfigure}
	\hfill
	\begin{subfigure}[b]{0.19\textwidth}
		\includegraphics[width=\textwidth]{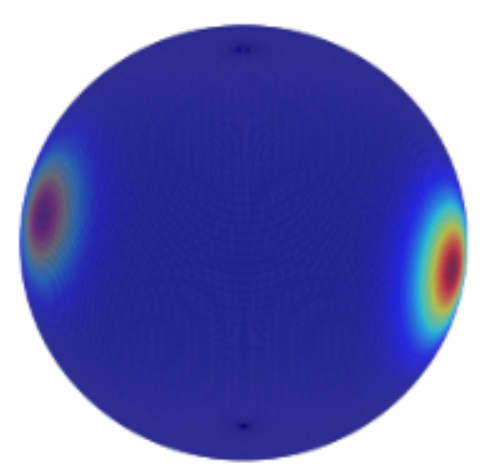}
		\caption{}
	\end{subfigure}
	\hfill
	\begin{subfigure}[b]{0.4\textwidth}
		\includegraphics[width=\textwidth]{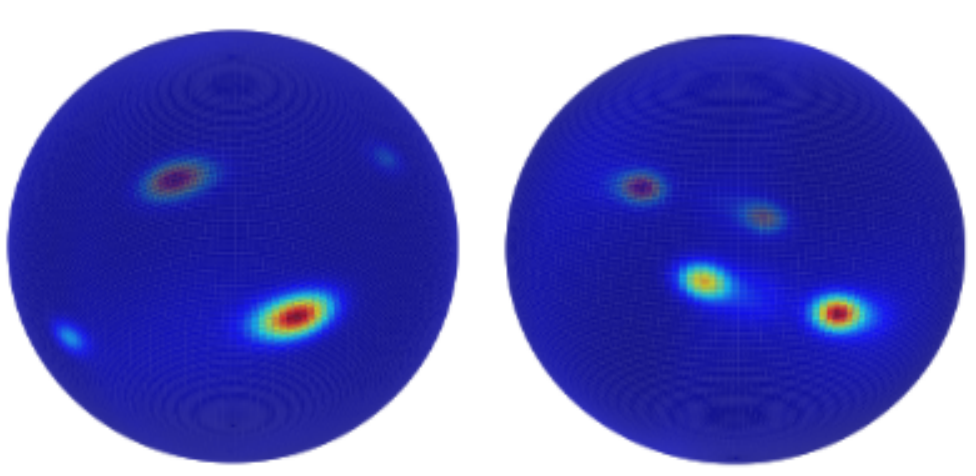}
		\caption{}
	\end{subfigure}
    \captionof{figure}{\footnotesize Bingham distributions plotted on the unit sphere: \textbf{(a)} low uncertainty, \textbf{(b)} higher uncertainty and \textbf{(c)} the mixtures of Ours-RWTA.}
	\label{fig:bingham_plot}
\end{minipage}
\end{figure}

%% file: tables/mhpANDmodes.tex
\begin{table}[h]%
\begin{minipage}[t]{.5\linewidth}
      \centering\small
  \caption{\small Ratio of correctly detected modes for various translational thresholds (in meters). A and B denote \textit{Blue Chairs} and \textit{Meeting Table} scenes.}%
  \vspace{2mm}
  \setlength{\tabcolsep}{3pt}
    \begin{tabular}{c|l|cccc}
    \multicolumn{1}{c}{Scene} & \multicolumn{1}{c}{Method} & 0.1 & 0.2 & 0.3 & 0.4 \\
    \midrule
    \multirow{2}[2]{*}{A} & MC-Dropout & 0.11  & 0.15  & 0.16  & 0.16 \\
          & Ours-RWTA & \textbf{0.36} & \textbf{0.79} & \textbf{0.80} & \textbf{0.80} \\
    \midrule
    \multirow{2}[1]{*}{B} & MC-Dropout & 0.04  & 0.07  & 0.09  & 0.11 \\
          & Ours-RWTA & \textbf{0.10} & \textbf{0.43} & \textbf{0.63} & \textbf{0.73} \\
          \bottomrule
    \end{tabular}%
  \label{table:mode_detection}
    \end{minipage}\hfill
\hspace{5mm}
    \begin{minipage}[t]{.45\linewidth}
      \caption{\small Comparison between different\\MHP variants, RWTA\cite{rupprecht2017learning} and EWTA\cite{makansi2019overcoming}, averaged over all scenes of our ambiguous real dataset.}
      \vspace{2mm}
		\setlength{\tabcolsep}{1.5pt}
			\begin{tabular}{lccc}
			
			    \multirow{2}{*}{Threshold} &  \multirow{2}{*}{\shortstack[c]{EWTA\\(k=50)}} & \multirow{2}{*}{\shortstack[c]{EWTA\\(k=25)}} & \multirow{2}{*}{\shortstack[c]{RWTA\\(k=1, used)}} \\
				\noalign{\smallskip}\noalign{\smallskip}\noalign{\smallskip}
				\midrule
				10$^\circ$ / 0.1m & 0.12 & 0.18 & \textbf{0.20}\\
				15$^\circ$ / 0.2m & 0.34 & 0.40 & \textbf{0.56}\\
				20$^\circ$ / 0.3m & 0.47 & 0.51 & \textbf{0.68}\\
				\noalign{\smallskip}
				\bottomrule
			\end{tabular}
		\label{table:ewta}
    \end{minipage} \vspace{-4mm}
\end{table}

%% file: tables/backbone_average.tex
\begin{table*}[t]
	\begin{center}
	\footnotesize
		\caption{Mean ratio of correct poses for different backbone networks on all scenes.}
		\resizebox{\textwidth}{!}{	
		\setlength{\tabcolsep}{6pt}
			\begin{tabular}{lcccccc}
			     & Threshold & PoseNet & Unimodal & Bingham-MDN & MC-Dropout & Ours-RWTA \\
				\noalign{\smallskip}
				\midrule
				& 10$^\circ$ / 0.1m & 0.15 & 0.12 & 0.08 & 0.15 & \textbf{0.20}\\
				ResNet-34 & 15$^\circ$ / 0.2m & 0.46 & 0.39 & 0.28 & 0.39 & \textbf{0.56}\\
				& 20$^\circ$ / 0.3m & 0.60 & 0.53 & 0.37 & 0.54 & \textbf{0.68}\\
				\midrule
				 & 10$^\circ$ / 0.1m& 0.15 & 0.16 & 0.09 & 0.15 & \textbf{0.19}\\
				ResNet-18 & 15$^\circ$ / 0.2m & 0.47 & 0.42 & 0.29 & 0.39 & \textbf{0.52}\\
				& 20$^\circ$ / 0.3m & 0.60 & 0.54 & 0.39 & 0.54 & \textbf{0.66}\\
				\midrule
				& 10$^\circ$ / 0.1m & \textbf{0.20} & 0.15 & 0.10 & 0.15 & \textbf{0.20}\\
				ResNet-50 & 15$^\circ$ / 0.2m & 0.49 & 0.36 & 0.30 & 0.40 & \textbf{0.55}\\
				& 20$^\circ$ / 0.3m & 0.62 & 0.53 & 0.38 & 0.53 & \textbf{0.69}\\
				\midrule
				& 10$^\circ$ / 0.1m & 0.11 & 0.10 & 0.11 & 0.08 & \textbf{0.18}\\
				Inception-v3 & 15$^\circ$ / 0.2m & 0.38 & 0.33 & 0.38 & 0.31 & \textbf{0.49}\\
				& 20$^\circ$ / 0.3m & 0.55 & 0.53 & 0.52 & 0.49 & \textbf{0.63}\\
				\noalign{\smallskip}
				\bottomrule
			\end{tabular}
		}
		\label{table:backbone}
	\end{center}\vspace{-5mm}
\end{table*}

%% file: conclusion.tex
\section{Conclusion}
\label{sec:conclude}
We have presented a novel method dealing with problems of direct camera pose regression in highly ambiguous environments where a unique solution to the 6DoF localization might be nonexistent.
Instead, we predict camera pose hypotheses as well as associated uncertainties that finally produce a mixture model. We use the Bingham distribution to model rotations and multivariate Gaussian distribution to obtain the position of a camera. In contrast to other methods like MC-Dropout~\cite{kendall2016modelling} or mixture density networks our training scheme is able to avoid mode collapse. Thus, we can obtain better mode predictions and improve upon the performance of camera pose regression methods in ambiguous environments while retaining the performance in non-ambiguous ones.

%% file: acknowledgements.tex
\small\noindent\textbf{Acknowledgements}: We would like to thank Johanna Wald for providing the Google Tango device. This project is supported by Bavaria California Technology Center (BaCaTeC), Stanford-Ford Alliance, NSF grant IIS-1763268, Vannevar Bush Faculty Fellowship, Samsung GRO program, the Stanford SAIL Toyota Research, and the PRIME programme of the German Academic Exchange Service (DAAD) with funds from the German Federal Ministry of Education and Research (BMBF).\vspace{-3mm}

%% file: appendix.tex
\setcounter{section}{0}
\renewcommand\thesection{\Alph{section}}
\newcommand{\suppsection}{\subsection}
\clearpage
\begin{flushleft}
\textbf{\large Appendix}
\end{flushleft}
\makeatletter
This appendix supplements our paper \textbf{6D Camera Relocalization in Ambiguous Scenes via Continuous Multimodal Inference}. In particular we present the following: (1) technical details on network architecture, training and modeling of translations, (2) more evaluations on synthetic data, (3) the used dataset, (4) error metrics, (5) additional quantitative (ablation) studies on uncertainty estimation, effect of the number of hypotheses on computational time, rotation parameterization and different means of assembling the Bingham matrix from the network output, (6) qualitative results on our real dataset.\vspace{-3mm}
\section{Modeling translations}\vspace{-1mm}
As described in the main paper we model translations using mixture density networks \cite{bishop1994mixture}. In more detail, for a sample input image $\Img\in \R^{W\times H \times 3}$, we obtain a predicted translation $\hat{\tb} \in\R^{c=3}$ from a neural network with parameters  $\bm{\Gamma}$. This prediction is set to the most likely value of a multivariate Gaussian distribution with covariance matrix
\begin{align}
    \mathbf{\Sigma} = \begin{bmatrix}
    \sigma_1^2 & & \\
    &  \ddots & \\
    &  & \sigma_{c}^2
    \end{bmatrix}_{c \times c},
\end{align}

\noindent where $\boldsymbol{\sigma}^2$ is predicted by our model.
As a result our model for a unimodal Gaussian is defined as:
\begin{equation}
\label{eq:gauss}
	p_{\bm{\Gamma}} (\tb \,|\, \Img) = \frac{\text{exp} (-\frac{1}{2}(\tb - \hat{\tb})^\top \mathbf{\Sigma}^{-1} (\tb - \hat{\tb}))}{(2\pi)^{c/2} |\mathbf{\Sigma}|^{1/2}} ,
\end{equation}

\noindent where $c=3$ and both $\hat{\tb}$ as well as $\mathbf{\Sigma}$ are trained by maximizing its log-likelihood.

Similar to forming a Bingham Mixture Model, we can equally compute a Gaussian Mixture Model with $K$ components and corresponding weights $\pi(\Img, \bm{\Gamma})$, such that $\sum_{j=1}^{K} \pi_j(\Img, \bm{\Gamma}) = 1$, to obtain a multi-modal solution. Again both $\hat{\tb}$ and $\mathbf{\Sigma}$ as well as $\pi(\Img, \bm{\Gamma})$ are learned by the network and trained by maximizing the log-likelihood of the mixture model. Note that, in this case, the components of $\hat{\tb}$ are assumed to be statistically independent within each distribution component. However, it has been shown that any density function can be approximated up to a certain error by a multivariate Gaussian mixture model with underlying kernel function as defined in \cref{eq:gauss} \cite{bishop1994mixture, mclachlan1988mixture}.
\vspace{-3mm}
\section{Network and training details}
We resize the input images to a height of 256 pixels and use random crops of size $224 \times 224$ for training. For testing we use the central crop of the image. As described in the main paper we use a ResNet-34~\cite{he2016deep} as our backbone network, which was pretrained on ImageNet \cite{ILSVRC15}, and remove the final classification layers. Fully-connected layers are then appended as specified in~\cref{table:layers}, where we output $K$ camera pose hypotheses, $\q$ and $\tb$, corresponding distribution parameters, $\mathbf{\Lambda}$ and $\mathbf{\Sigma}$, as well as shared mixture weights $\pi(\Img, \bm{\Gamma})$. In case of our single component and Bingham-MDN models we use a softmax activation function, such that $\sum_{j=1}^{K} \pi_j(\Img, \bm{\Gamma}) = 1$ holds true. In our MHP version, we first apply a ReLU activation function, that, during training, is passed to a cross-entropy loss function. Once trained, we again apply a softmax on the final weights to form a valid mixture model. 
\input{supplementary/layers}

\begin{figure*}[b]
	\centering
	\includegraphics[width=\textwidth]{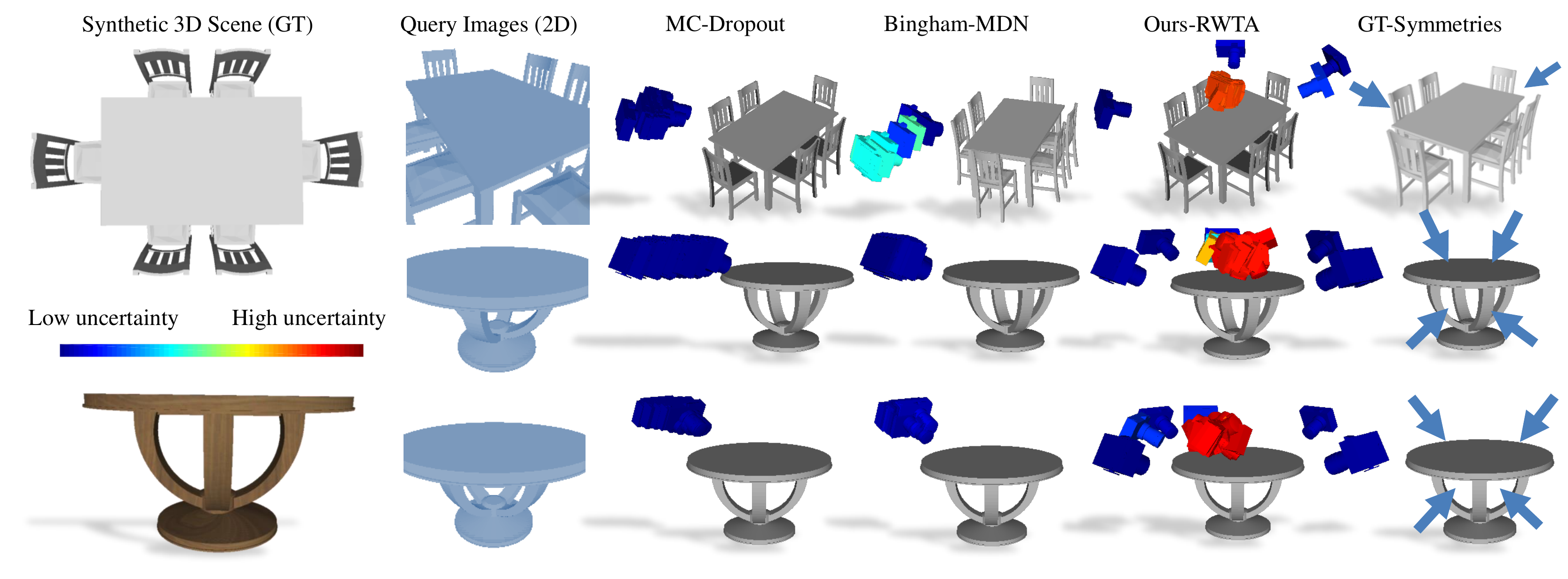}
	\caption{Additional qualitative results of our synthetically created dataset. If available, camera poses are colored by their uncertainty.}
	\label{fig:synthSup}
\end{figure*}
\begin{figure*}[h]
	\centering
	\includegraphics[width=\textwidth]{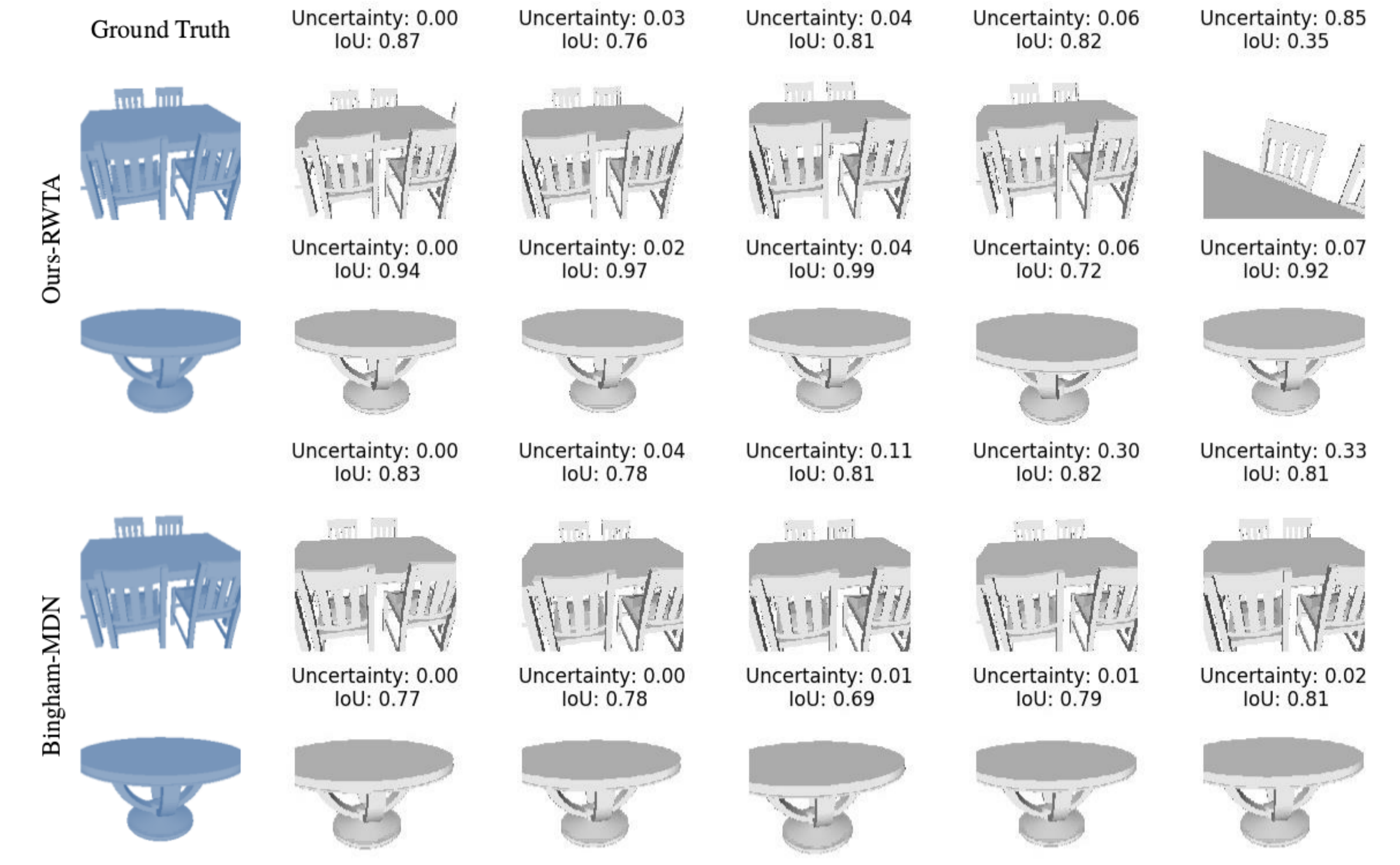}
	\caption{Renderings of the top five camera pose hypotheses according to their uncertainty values for our Bingham-MDN and MHP version, Ours-RWTA. Further we show the corresponding ground truth query images as well as the intersection over union of the ground truth and predicted renderings. }
	\label{fig:hypotheses}
\end{figure*}
\begin{figure*}[t]
	\centering
	\begin{subfigure}[b]{0.25\textwidth}
		\includegraphics[width=\textwidth]{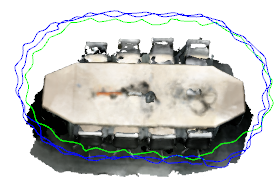}
	\end{subfigure}
	\begin{subfigure}[b]{0.25\textwidth}
		\includegraphics[width=\textwidth]{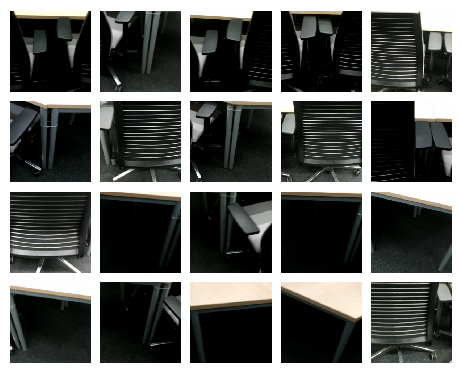}
	\end{subfigure}	
	\begin{subfigure}[b]{0.2\textwidth}
		\includegraphics[width=\textwidth]{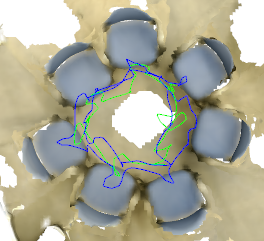}
	\end{subfigure}
	\begin{subfigure}[b]{0.25\textwidth}
		\includegraphics[width=\textwidth]{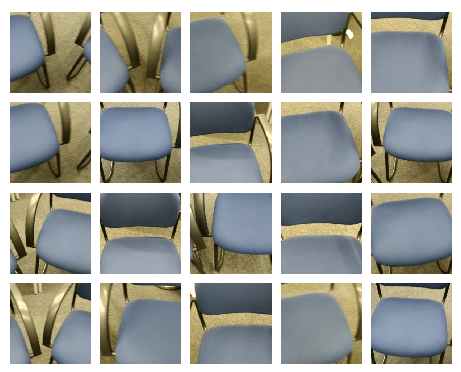}
	\end{subfigure}
	\begin{subfigure}[b]{0.25\textwidth}
		\includegraphics[width=\textwidth]{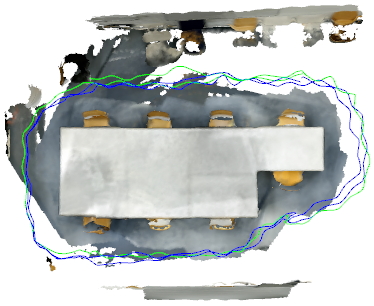}
	\end{subfigure}
	\begin{subfigure}[b]{0.25\textwidth}
		\includegraphics[width=\textwidth]{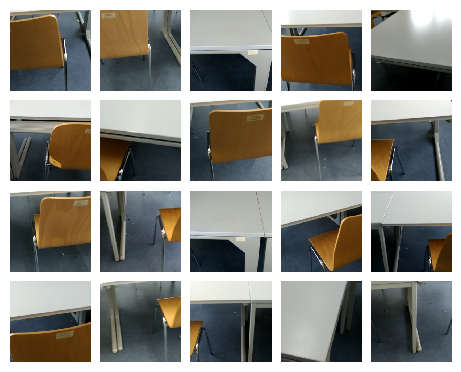}
	\end{subfigure}
	\begin{subfigure}[b]{0.2\textwidth}
		\includegraphics[width=\textwidth]{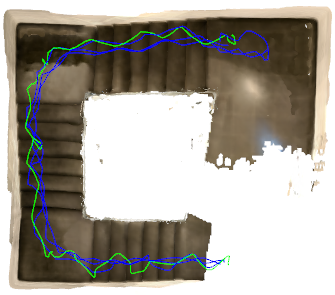}
	\end{subfigure}
	\begin{subfigure}[b]{0.25\textwidth}
		\includegraphics[width=\textwidth]{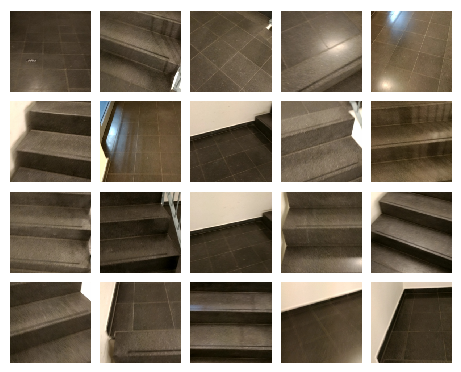}
	\end{subfigure}
	\caption{Ground truth training (\textcolor{blue}{blue}) and test (\textcolor{green}{green}) trajectories of our ambiguous scenes and example RGB images.}
	\label{fig:dataset}
\end{figure*}
For our single component model we set $K=1$, otherwise we use $K=50$ hypotheses. For all models we train with an initial learning rate of $1e^{-4}$.

\section{Evaluation on synthetic scenes}
\begin{figure}
	\centering
	\begin{subfigure}[t]{0.25\textwidth}
		\includegraphics[width=\textwidth]{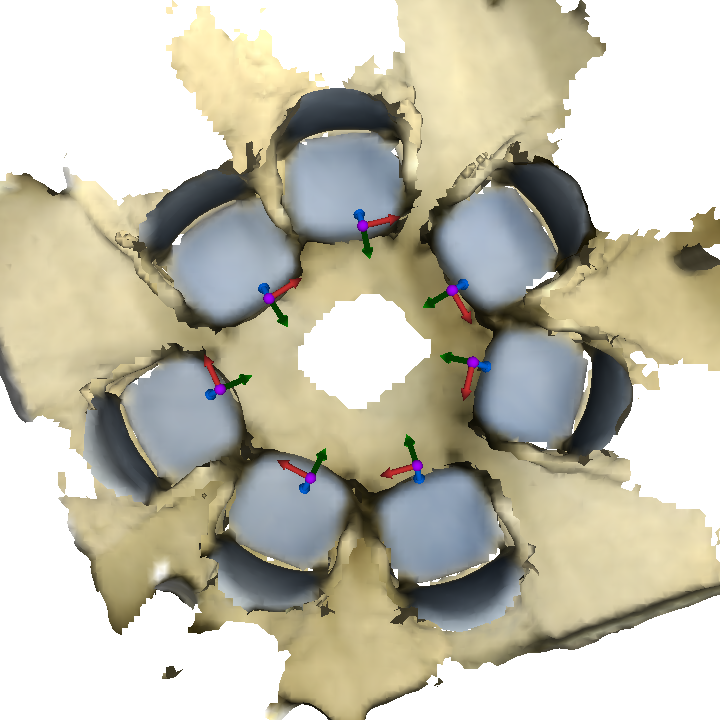}
	\end{subfigure}
	\begin{subfigure}[t]{0.25\textwidth}
		\includegraphics[width=\textwidth]{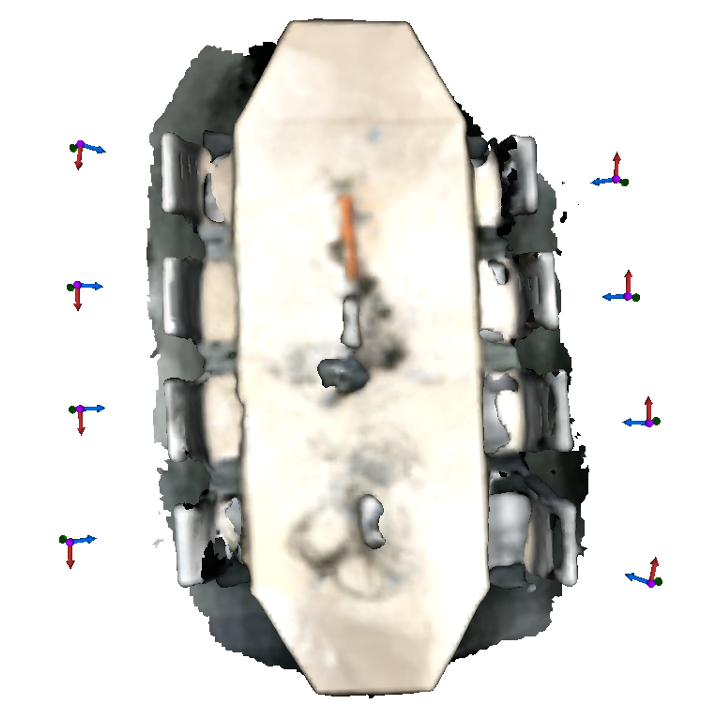}
		\label{fig:modes}
	\end{subfigure}
	\hfill
\begin{subfigure}[t]{0.48\textwidth}
	\includegraphics[width=\textwidth]{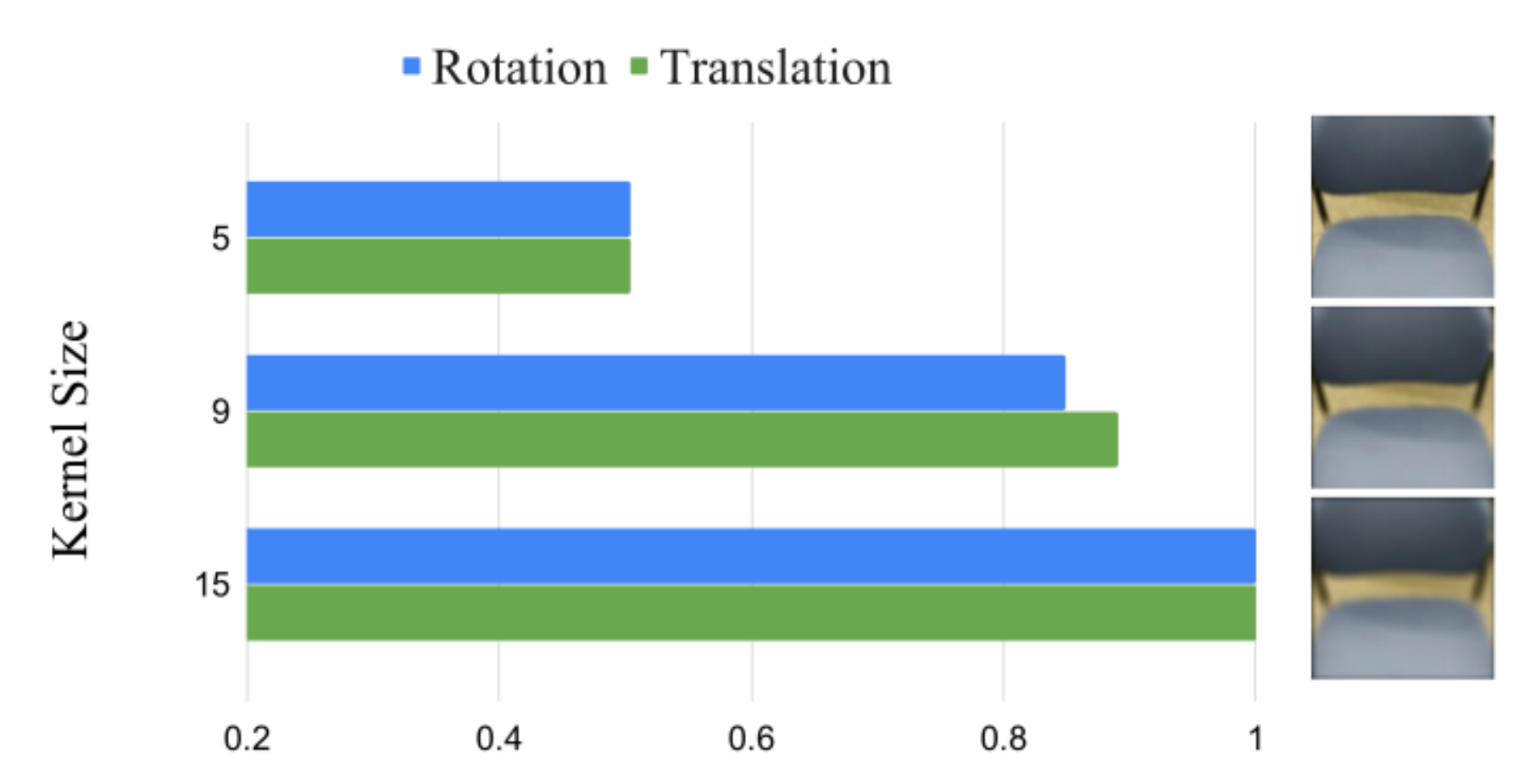}
	\label{fig:blur}
	\end{subfigure}
	\caption{\textbf{(left)} Estimated ground truth modes in \textit{Blue Chairs} and \textit{Meeting Table} scenes, which we use to evaluate our model's mode detection performance and diversity of predictions. \textbf{(right)} Change in uncertainty prediction in the presence of increasing image blur. For varying kernel sizes of a Gaussian filter used to blur the input images, we compute the average uncertainty over all images obtained from the predictions of our model. Reported here are the normalized values.}
	\label{fig:modesAndBlur}%
\end{figure}

We now show in~\cref{fig:synthSup} different query images and localization results on the synthetic scenes provided in the paper. The superiority of our approach is consistent across different viewpoints. 
We also provide additional quantitative and qualitative results on our synthetic dataset. For this aim, we render the objects/scenes from the predicted camera poses of our methods in~\cref{fig:hypotheses}. There, we show the most certain predictions sorted according to the entropy of the resulting Bingham and Gaussian distributions.

Last but not least, we compute the intersection over union (IoU) with the renderings obtained from the ground truth camera poses. Considering the hypothesis with the highest weight as the single best prediction, on average our Bingham-MDN reaches an IoU of 0.62, whereas our MHP distribution model, \textit{Ours-RWTA}, achieves 0.88. 
\vspace{-3mm}
\section{Details on the acquisition of real ambiguous dataset}
Besides our synthetically created dataset, we captured a highly ambiguous real dataset, consisting of five scenes using Google Tango~\cite{eitan2016}.~\cref{fig:dataset} shows ground truth training and testing camera trajectories, plotted with Open3D~\cite{Zhou2018}, as well as example batch images we acquired for our ambiguous scene dataset. 
The resolution of the captured RGB images is $540 \times 960$ and the spatial extent of our scenes can be found in \cref{tab:dimensions}.
\input{supplementary/spatial}
Further, for each image in the \textit{Blue Chairs} and \textit{Meeting Table} scenes, we obtain a ground truth estimate by training an autoencoder on reconstructing the input images and using the resulting feature descriptors to obtain the nearest neighbor camera poses. Then we cluster the resulting camera poses using a Riemannian Mean Shift algorithm~\cite{subbarao2009nonlinear} and use the centroids of the resulting clusters as "ground truth" modes. We visually verify the results. The autoencoder we use to compute said features contains a ResNet-34 encoder, followed by subsequent deconvolutions with batch normalization and ReLU activation as the decoder. It is trained with an $l_2$ reconstruction loss for 300 epochs using the Adam Optimizer~\cite{kingma2014adam} with a learning rate of $1e^{-3}$ and a batch size of 20 images. Examples of the obtained ground truth modes can be found in~\cref{fig:modesAndBlur} (left).

\vspace{-1mm}\section{Error Metrics}
Given a ground truth camera pose, consisting of a rotation, represented by a quaternion $\q$, and its translation, $\tb$, we evaluate the performance of our models with respect to the accuracy of the predicted camera poses by computing the recall of ours and the baseline models. We consider a camera pose estimate to be correct if both rotation and translation are below a pre-defined threshold and compute the angular error between GT, $\q$, and predicted quaternion, $\hat{\q}$, as 
\begin{equation}
    d_q(\q,\hat{\q}) = 2\arccos(|\q \circ \hat{\q}|).
\end{equation}
For translations we use the norm of the difference between GT $\tb$, and predicted translation $\hat{\tb}$: $
    d_t(\tb,\hat{\tb}) = \|\tb - \hat{\tb}\|_2$ to compute the error in position of the camera.
\input{supplementary/time}
\section{Further Ablation Studies}
\subsection{Uncertainty evaluation}
Due to fast camera movements, motion blur easily arises in camera localization applications and is one factor that can lead to poor localization performance. As a first step in handling such problems, additional information in the form of uncertainty predictions could aid in detecting such events. Therefore, to evaluate how our model performs in the presence of noise, we use our single component model, i.e. $K=1$, trained on the original input images, and blur the RGB images to evaluate the change in uncertainty prediction of the model. Ideally, with increasing image blur, we would expect our model to be less certain in its predictions. To ablate on this, we apply a Gaussian Filter to the input images, with varying kernel sizes, and report the change in uncertainty prediction in~\cref{fig:modesAndBlur} (right) on the blurred images. We use the entropy over each image to obtain a measure of uncertainty and compute the mean over our dataset images. For visualization, we show the normalized values. An increase in uncertainty could be clearly observed with growing kernel size and thus highly blurred images.
\subsection{Number of Hypotheses and Computational Times}
Incorporating our method into an existing regression model, simply leads to a change in the last fully-connected layers of the network. We extend the last layer to output an additional $(K-1) \cdot 4$ and $(K-1) \cdot 3$ parameters for predicting the camera pose, as well as overall $6 \cdot K$ for uncertainty prediction of both rotation and translation. Further, we incorporate extra layers for the mixture coefficients as described in \cref{table:layers}. We run our model on a 8GB NVIDIA GeForce GTX 1080 graphics card and report the inference time of our network with respect to $K$ in \cref{tab:time}. In comparison to a direct regression method our model with $K=50$ incurs a negligible computational overhead around $1ms$.
 \begin{figure*}[t]
	\centering
	\includegraphics[width=\textwidth]{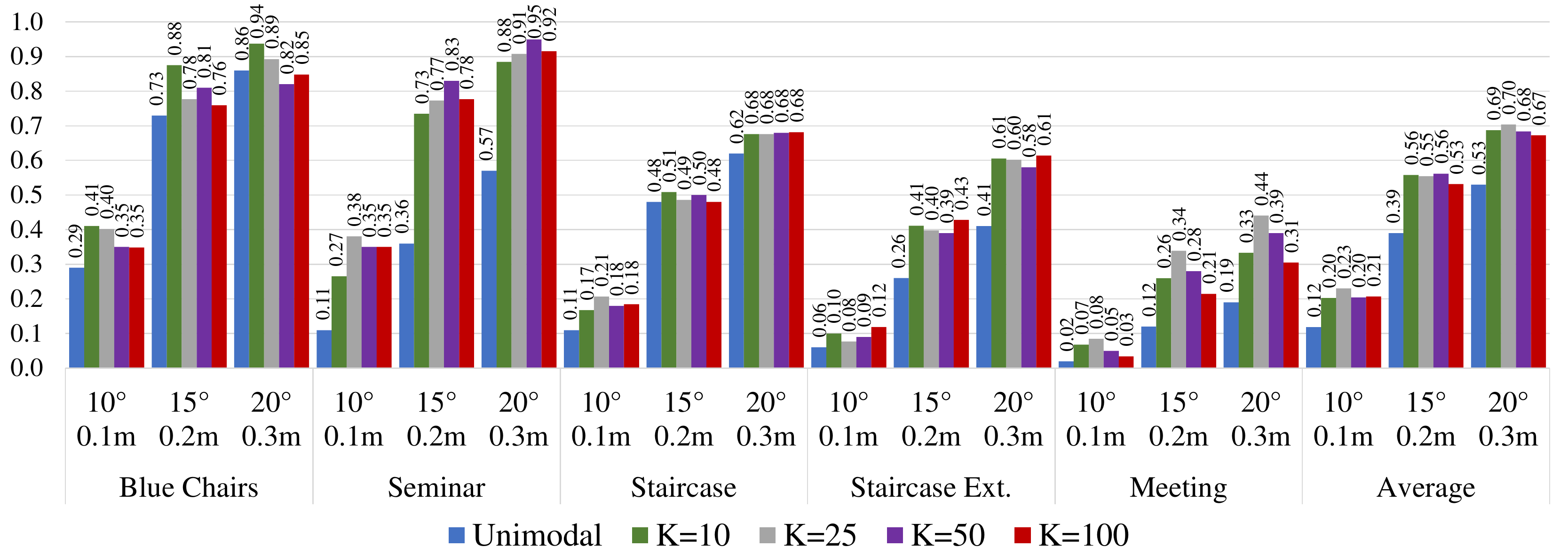}
	\caption{Influence of the number of hypotheses, i.e. parameter $K$, on the performance of our method, Ours-RWTA.}
	\label{fig:numK}%
\end{figure*}

Further, we evaluate the effect of hyperparameter $K$, i.e. the number of hypothesis to be regressed, for our proposed method. Based on the results, which are summarized in \cref{fig:numK}, we suspect the optimal number of hypotheses to be dependant on the spatial extent of the scene and on the ambiguities contained in them. However, due to the increased complexity of the model as well as instability issues during training, we observed a drop in performance with high increase of the number of hypotheses.
\input{supplementary/continuous}

\subsection{Rotation parameterization}
The best choice of rotation parameterization for training deep learning models is an open question. PoseNet~\cite{kendall2015posenet} proposed to use quaternions due to the ease of normalization. The ambiguities can be resolved by mapping the predictions to one hemisphere. MapNet~\cite{brahmbhatt2018geometry} further showed improvements in using the axis angle representation. Recently it has been shown that any representation with four or less degrees of freedom suffers from discontinuities in mapping to $SO(3)$. This might harm the performance of deep learning models. Instead,~\cite{zhou2019continuity} proposed a continuous 6D or 5D representation. We ablate in this context by mapping all predictions to the proposed 6D representation and model them using a GMM, similar to a MDN, but treating rotation and translation separately. Therefore for each camera pose, in total we have $9 \cdot 2$ parameters to regress, plus mixture coefficients.~\cref{table:6dambigious} shows our results, where 'Geo + L1' refers to a direct regression using the geodesic loss proposed in \cite{zhou2019continuity} and an $l_1$ loss on the translation. When using the proposed 6D representation, we found either improvements or similar performance to their quaternion counterparts. However, overall our 9D-Ours-RWTA remains the most promising model. In terms of Oracle Error MC-Dropout sometimes outperforms our method. This comes from the fact that MC-Dropout mostly predicts multiple hypothesis around one mode, which if this mode is relatively close to the ground truth one, results in a high Oracle. 9D-Ours-RWTA predicts diverse hypothesis, but not multiple versions of the same mode. However it shows much better performance in predicting the correct mode than MC-Dropout.
 \input{supplementary/constructionV}

\subsection{Ablation studies for constructing $\V$} Alternatively to the method proposed in the main paper, Gram-Schmidt can be used to compute an orthonormal matrix $\V$ from a given matrix $\mathbf{M} \in \R^{d \times d}$, where the column vectors $\mathbf{v_i}$ of $\mathbf{V}$ are computed from the column vectors $\mathbf{m_i}$ as follows
\begin{equation}
		\mathbf{\hat{v}_i} = \mathbf{m}_i - \sum_{k=1}^{i-1} \langle \mathbf{v}_k,\mathbf{m}_i \rangle \cdot \mathbf{v}_k~,\text{where}~
		\mathbf{v}_i = \frac{\mathbf{\hat{v}}_i}{\|\mathbf{\hat{v}}_i\|}.
\end{equation}
Note that in the GS procedure, we predict 16 values for $\V$ and use GS to project onto the orthonormal matrices. Yet the degrees of freedom of $\V$ is much less. For instance the matrix scheme of~\cite{birdal2018bayesian} uses only four. As an ablation, we propose another way to construct $\V$ using the \textit{Cayley transform}~\cite{cayley1846quelques} as follows: Given a vector $\q$ (not necessarily with unit norm), we compute $\mathbf{V}$ as
\begin{equation}
	\V = (\Eye_{d \times d} - \mathbf{S})^{-1} (\Eye_{d \times d} + \mathbf{S}),
\end{equation}
where $\Eye_{d \times d}$ is the identity matrix and
\begin{align}
\label{eq:S}
\mathbf{S}(\q) \triangleq 
\begin{bmatrix}
\phantom{-}0       & -q_1 	        & q_4 	        &  -q_3 \\
\phantom{-}q_1     & \phantom{-}0 	& q_3 	        &  \phantom{-}q_2\\
-q_4               & -q_3 	        & \phantom{-}0 	&  -q_1\\
\phantom{-}q_3     & -q_2 	        & q_1	        &  \phantom{-}0
\end{bmatrix}
\end{align}
a skew-symmetric matrix parameterized by $\q$. We compare between the proposed method used in the paper and these two alternatives, GS orthonormalization and the construction using skew-symmetric matrices. The results can be found in \cref{table:V}. In comparison to GS the remaining methods only require four parameters to be estimated instead of the 16 entries of the matrix $\V$. For our unimodal as well as multimodal MDN we found the Skew-Symmetric construction to outperform both Gram-Schmidt (GS) and the employed method of Birdal~\etal~\cite{birdal2018bayesian}. However, for our method, \emph{Ours-RWTA}, the latter~\cite{birdal2018bayesian} performs the best. Additionally it achieves overall the best performance in comparison to the remaining methods and constructions.

\paragraph{\textbf{Expressive power of $\V$}}
In this section as well as in the main paper we have presented a variety of ways to establish $\V(\q)$. These methods range from regressing $16$ parameters (full flexibility) to regressing only $4$ (less flexibility but also a smaller parameter space). This raises an interesting trade-off on the expressiveness of $\V$ and the performance of the neural network \ie how many parameters would \emph{suffice} to capture all the necessary Bingham distributions? This remains to be an open question as, like many others, Birdal~\etal~\cite{birdal2018bayesian} (our choice of construction) did not provide an analysis on the extent of sufficiency. Nevertheless, we would like to point out that once $\V$ is chosen to be a particular frame, it can explain other orthogonal bases as a linear combination. This introduces certain degree of expressive power (albeit quantized), which we have empirically found to be sufficient compared to other potentially over-parameterized schemes. Besides, the method of Birdal~\etal~\cite{birdal2018bayesian} is computationally the cheapest.

\section{Additional qualitative results}
\begin{figure*}[h]
	\centering
	\includegraphics[width=\textwidth]{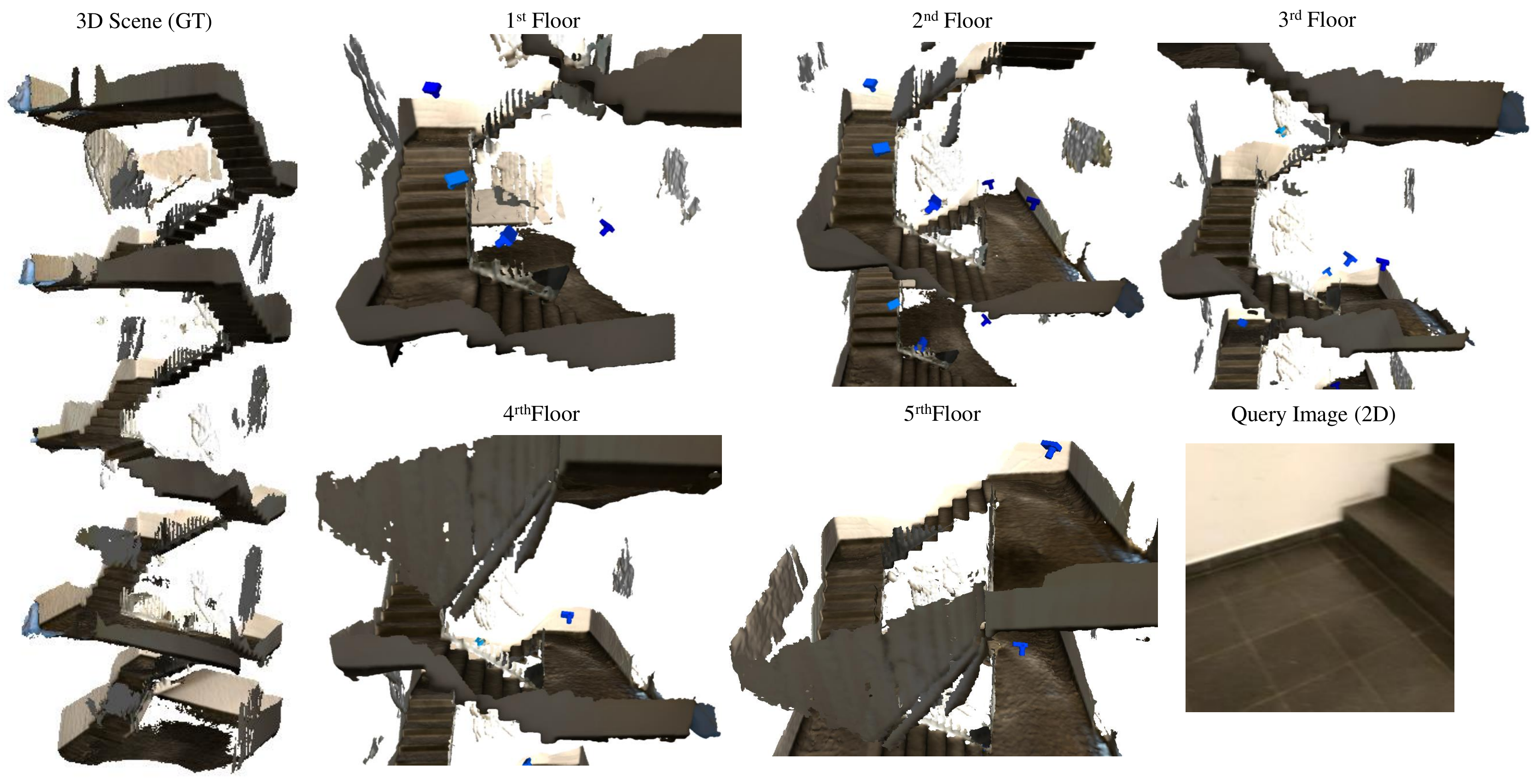}
	\caption{Qualitative results of our model, Ours-RWTA, on the \textit{Staircase Extended} scene.}\vspace{-4mm}
	\label{fig:real_stairs}
\end{figure*}

\begin{figure*}[t]
	\centering
	\includegraphics[width=\textwidth]{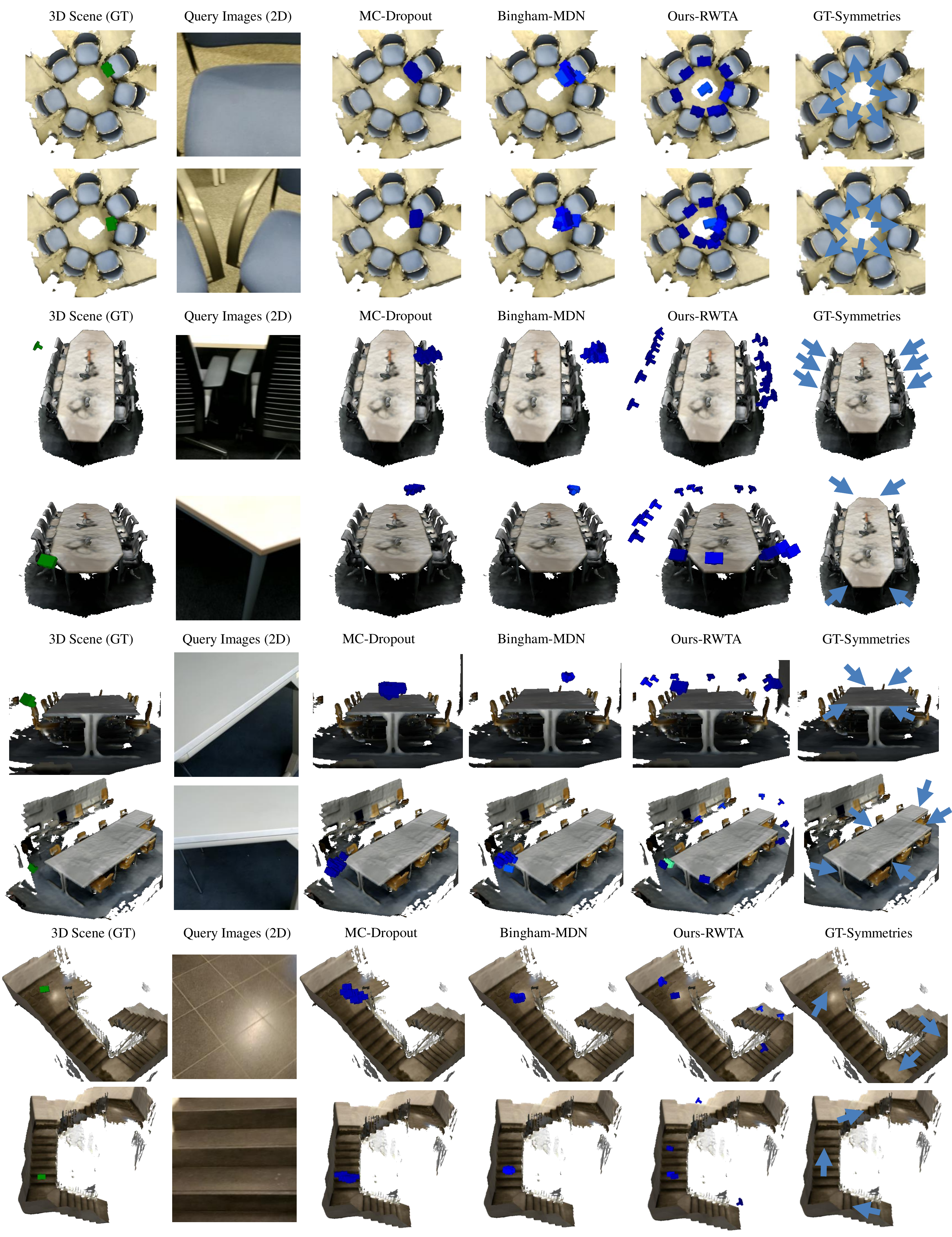}
	\caption{Additional qualitative results of our ambiguous scenes dataset. We show the ground truth camera pose, query images and resulting camera pose predictions. Both MC-Dropout and our Bingham-MDN suffer from mode collapse, whereas our MHP-based model, Ours-RWTA, predicts diverse hypotheses covering all possible modes.}
	\label{fig:real_blue}
\end{figure*}

 Further, we provide more qualitative results from different query images on all scenes of our ambiguous dataset in~\cref{fig:real_stairs} and~\cref{fig:real_blue}.

%% file: supplementary/layers.tex
\begin{table}[t]
	\begin{center}
		\caption{Layer specifications. We report the dimensionality of the input feature vector, $N_{in}$, resulting output feature vector, $N_{out}$, whether or not batch normalization (BN) is used and the activation function for each layer.}
		\vspace{5pt}
		\resizebox{0.4\textwidth}{!}{		
			\begin{tabular}{lcccc}
				& $N_{in}$ & $N_{out}$ & BN & activation\\
				\noalign{\smallskip}
				\midrule
				\noalign{\smallskip}
				 $\q$ & 2048 & $K$*4 & no & none\\
				 $\tb$ & 2048 & $K$*3 & no & none\\
 				 $\mathbf{\Lambda}$ & 2048 & $K$*3 & no & softplus\\
				 $\mathbf{\Sigma}$ & 2048 & $K$*3 & no & softplus\\
				 $\boldsymbol{\pi}$ & 2048 & 1024 & yes & ReLU\\
 				 & 1024 & 512 & yes & ReLU\\
				 & 512 & $K$ & yes & ReLU / softmax\\
				\bottomrule
			\end{tabular}
		}
		\label{table:layers}\vspace{-6mm}
	\end{center}
\end{table}

%% file: supplementary/spatial.tex
 \begin{table}[t]
	\begin{center}
		\setlength{\tabcolsep}{5pt}
		\caption{Spatial Extent of our scenes in meters.}
		\vspace{5pt}
		\resizebox{0.9\textwidth}{!}{		
			\begin{tabular}{ccccc}
                Blue Chairs & Meeting Table  & Staircase & Staircase Ext. & Seminar Room\\
				\noalign{\smallskip}
				\midrule
				\noalign{\smallskip}
               $ 5\times 4.6\times 1.3$ & $4.3\times 5.8\times 1.4$ & $4.9\times 4.4\times 5.1$ & $5.6\times 5.2\times 16.6$ & $5.3\times 7.8\times 2.6$\\
				\bottomrule
			\end{tabular}
		}
		\label{tab:dimensions}
	\end{center}%
\end{table}

%% file: supplementary/time.tex
 \begin{table}[b!]
    \centering%
    \caption{Inference time of our method with respect to the number of hypotheses.}
    \vspace{5pt}
		\resizebox{0.5\textwidth}{!}{		
			\begin{tabular}{ccccc}
                PoseNet & $K=1$ & $K=50$ & $K=200$ & $K=500$\\
				\noalign{\smallskip}
				\midrule
				\noalign{\smallskip}
                $7.23ms$ & $7.27ms$ & $8.11ms$ & $8.19ms$ & $8.74ms$\\
				\bottomrule
			\end{tabular}
		}
		\label{tab:time}%
\end{table}

%% file: supplementary/continuous.tex
\begin{table*}[t]
	\begin{center}
		\caption{Ratio of correct poses when using the continuous 6D representation of~\cite{zhou2019continuity} to model rotations instead of a Bingham distribution on the quaternion.}
		\vspace{5pt}
		\resizebox{0.99\textwidth}{!}{		
			\begin{tabular}{lcccccc|cc}
				& Threshold & Geo+L1 & Uni. & MDN & \vtop{\hbox{\strut ~~~MC- }\hbox{\strut Dropout}} & \vtop{\hbox{\strut 9D-Ours- }\hbox{\strut ~RWTA}} & \vtop{\hbox{\strut MC-Dropout }\hbox{\strut ~~~~Oracle}} & \vtop{\hbox{\strut 9D-Ours-RWTA }\hbox{\strut ~~~~Oracle}} \\
				\noalign{\smallskip}
				\midrule
				\noalign{\smallskip}
				& 10$^\circ$ / 0.1m & 0.41 & \textbf{0.48} & 0.01 & 0.26 & 0.38 & \textbf{0.58} & 0.44\\
				Blue Chairs &15$^\circ$ / 0.2m & \textbf{0.90} & 0.89 & 0.14 & 0.83 & 0.81 & 0.91 & \textbf{0.96}\\
				&20$^\circ$ / 0.3m & \textbf{0.96} & 0.92 & 0.23 & 0.91 & 0.84 & 0.94 & \textbf{1.0}\\
				\midrule
				&10$^\circ$ / 0.1m & 0.03 & 0.03 & 0.02 & 0.02 & \textbf{0.06} & \textbf{0.09} & 0.08\\
				Meeting Table &15$^\circ$ / 0.2m & 0.16 & 0.16 & 0.11 & 0.13 & \textbf{0.29} & 0.24 & \textbf{0.47}\\
				&20$^\circ$ / 0.3m & 0.22 & 0.23 & 0.14 & 0.21 & \textbf{0.38} & 0.32 & \textbf{0.71}\\
				\midrule
				&10$^\circ$ / 0.1m & 0.17 & \textbf{0.19} & 0.12 & 0.12 & 0.18 & \textbf{0.27} & 0.21\\
				Staircase &15$^\circ$ / 0.2m & 0.46 & \textbf{0.51} & 0.36 & 0.36 & 0.44 & \textbf{0.56} & 0.54\\
				&20$^\circ$ / 0.3m & 0.62 & \textbf{0.67} & 0.47 & 0.56 & 0.56 & 0.70 & \textbf{0.71}\\
				\midrule
				&10$^\circ$ / 0.1m & 0.07 & 0.01 & 0.01 & 0.04 & \textbf{0.08} & \textbf{0.15} & 0.09\\
				Staircase Extended &15$^\circ$ / 0.2m & 0.30 & 0.06 & 0.09 & 0.18 & \textbf{0.35} & \textbf{0.40} & 0.38\\
				&20$^\circ$ / 0.3m & 0.48 & 0.13 & 0.14 & 0.36 & \textbf{0.55} & 0.59 & \textbf{0.62}\\
				\midrule
				&10$^\circ$ / 0.1m & \textbf{0.34} & 0.24 & 0.30 & 0.21 & \textbf{0.34} & \textbf{0.45} &  0.34\\
				Seminar Room&15$^\circ$ / 0.2m & 0.74 & 0.63 & 0.65 & 0.65 & \textbf{0.76} & \textbf{0.84} & 0.77\\
				&20$^\circ$ / 0.3m & 0.84 & 0.79 & 0.76 & 0.82 & \textbf{0.88} & \textbf{0.9} & 0.88\\
				\midrule\midrule
				&10$^\circ$ / 0.1m & 0.20 & 0.19 & 0.09 & 0.13 & \textbf{0.21} & \textbf{0.31} & 0.23\\
				Average &15$^\circ$ / 0.2m & 0.51 & 0.45 & 0.27 & 0.43 & \textbf{0.53} & 0.59 & \textbf{0.62}\\
				&20$^\circ$ / 0.3m & 0.63 & 0.55 & 0.35 & 0.57 & \textbf{0.64} & 0.69 & \textbf{0.78}\\
				\bottomrule
			\end{tabular}
		}
		\label{table:6dambigious}
	\end{center}%
\end{table*}

%% file: supplementary/constructionV.tex
\begin{table*}[t]
	\begin{center}
		\caption{Ratio of correct poses for several thresholds of Gram-Schmidt (\textit{G}), Skew-Symmetric (\textit{S}) and Birdal~\etal~\cite{birdal2018bayesian} (\textit{B}) methods to construct $\V$.}
		\vspace{5pt}
		\resizebox{0.99\textwidth}{!}{	
		\setlength{\tabcolsep}{5.0pt}
			\begin{tabular}{lcccc}
				 &  & Unimodal & Bingham-MDN & Ours-RWTA \\
				 \cmidrule(l){3-5}
				 & Threshold & G / S / B & G / S / B & G / S / B \\
				\noalign{\smallskip}
				\midrule
				\noalign{\smallskip}
				& 10$^\circ$ / 0.1m & 0.24 / 0.23 / 0.29 & 0.04 / 0.17 / 0.24 & 0.30 / 0.12 / 0.35\\
				Blue Chairs (A) & 15$^\circ$ / 0.2m & 0.63 / 0.58 / 0.73 & 0.15 / 0.49 / 0.75 & 0.73 / 0.39 / 0.81\\
				&20$^\circ$ / 0.3m & 0.76 / 0.73 / 0.86 & 0.18 / 0.59 / 0.80 & 0.79 / 0.43 / 0.82\\
				\midrule
				&10$^\circ$ / 0.1m & 0.02 / 0.07 / 0.02 & 0.04 / 0.01 / 0.01 & 0.04 / 0.09 / 0.05\\
				Meeting Table (B) & 15$^\circ$ / 0.2m & 0.16 / 0.20 / 0.12 & 0.18 / 0.14 / 0.07 & 0.12 / 0.23 / 0.28\\
				&20$^\circ$ / 0.3m & 0.24 / 0.25 / 0.19 & 0.21 / 0.24 / 0.10 & 0.18 / 0.27 / 0.39\\
				\midrule
				&10$^\circ$ / 0.1m & 0.17 / 0.16 / 0.11 & 0.21 / 0.16 / 0.04 & 0.17 / 0.14 / 0.18\\
				Staircase (C) &15$^\circ$ / 0.2m & 0.46 / 0.51 / 0.62 & 0.43 / 0.37 / 0.15 & 0.46 / 0.42 / 0.50\\
				&20$^\circ$ / 0.3m & 0.62 / 0.64 / 0.62 & 0.60 / 0.49 / 0.25 & 0.60 / 0.62 / 0.68\\
				\midrule
				&10$^\circ$ / 0.1m & 0.04 / 0.04 / 0.06 & 0.04 / 0.07 / 0.06 & 0.05 / 0.06 / 0.09\\
				Staircase Extended (D) & 15$^\circ$ / 0.2m & 0.16 / 0.16 / 0.26 & 0.19 / 0.29 / 0.21 & 0.23 / 0.26 / 0.39\\
				&20$^\circ$ / 0.3m & 0.27 / 0.27 / 0.41 & 0.31 / 0.41 / 0.32 & 0.34 / 0.36 / 0.58\\
				\midrule
				&10$^\circ$ / 0.1m & 0.27 / 0.33 / 0.06 & 0.30 / 0.35 / 0.06 & 0.15 / 0.28 / 0.35\\
				Seminar Room (E)& 15$^\circ$ / 0.2m & 0.69 / 0.69 / 0.23 & 0.56 / 0.59 / 0.23 & 0.47 / 0.70 / 0.83\\
				&20$^\circ$ / 0.3m & 0.82 / 0.80 / 0.40 & 0.64 / 0.70 / 0.40 & 0.58 / 0.79 / 0.95\\
				\midrule\midrule
				&10$^\circ$ / 0.1m & 0.15 / \textbf{0.16} / 0.11 & \textbf{0.13} / \textbf{0.13} / 0.08 & 0.14 / 0.14 / \textbf{0.20}\\
				Average &15$^\circ$ / 0.2m & 0.42 / \textbf{0.43} / 0.36 & 0.30 / \textbf{0.38} / 0.28 & 0.40 / 0.40 / \textbf{0.56}\\
				&20$^\circ$ / 0.3m & \textbf{0.54} / \textbf{0.54} / 0.50 & 0.39 / \textbf{0.49} / 0.37 & 0.50 / 0.49 / \textbf{0.68}\\
				\bottomrule
			\end{tabular}
		}
		\label{table:V}
	\end{center}%
\end{table*}

%% file: ArxivSubmission.bbl
\begin{thebibliography}{10}
\providecommand{\url}[1]{\texttt{#1}}
\providecommand{\urlprefix}{URL }
\providecommand{\doi}[1]{https://doi.org/#1}

\bibitem{arun2018probabilistic}
Arun~Srivatsan, R., Xu, M., Zevallos, N., Choset, H.: Probabilistic pose
  estimation using a bingham distribution-based linear filter. The
  International Journal of Robotics Research  \textbf{37}(13-14),  1610--1631
  (2018)

\bibitem{balntas2018relocnet}
Balntas, V., Li, S., Prisacariu, V.: Relocnet: Continuous metric learning
  relocalisation using neural nets. In: Proceedings of the European Conference
  on Computer Vision (ECCV). pp. 751--767 (2018)

\bibitem{barfoot2014associating}
Barfoot, T.D., Furgale, P.T.: Associating uncertainty with three-dimensional
  poses for use in estimation problems. IEEE Transactions on Robotics
  \textbf{30}(3) (2014)

\bibitem{Bingham1974}
Bingham, C.: An antipodally symmetric distribution on the sphere. The Annals of
  Statistics pp. 1201--1225 (1974)

\bibitem{birdal2020measure}
Birdal, T., Arbel, M., Şimşekli, U., Guibas, L.: Synchronizing probability
  measures on rotations via optimal transport. In: Proceedings of the IEEE
  Conference on Computer Vision and Pattern Recognition (2020)

\bibitem{birdal2016online}
Birdal, T., Bala, E., Eren, T., Ilic, S.: Online inspection of 3d parts via a
  locally overlapping camera network. In: 2016 IEEE Winter Conference on
  Applications of Computer Vision (WACV). pp. 1--10. IEEE (2016)

\bibitem{birdal2019probabilistic}
Birdal, T., Simsekli, U.: Probabilistic permutation synchronization using the
  riemannian structure of the birkhoff polytope. In: Proceedings of the IEEE
  Conference on Computer Vision and Pattern Recognition. pp. 11105--11116
  (2019)

\bibitem{birdal2018bayesian}
Birdal, T., Simsekli, U., Eken, M.O., Ilic, S.: Bayesian pose graph
  optimization via bingham distributions and tempered geodesic mcmc. In:
  Advances in Neural Information Processing Systems. pp. 308--319 (2018)

\bibitem{bishop1994mixture}
Bishop, C.M.: Mixture density networks  (1994)

\bibitem{bourmaud2015continuous}
Bourmaud, G., M{\'e}gret, R., Arnaudon, M., Giremus, A.: Continuous-discrete
  extended kalman filter on matrix lie groups using concentrated gaussian
  distributions. Journal of Mathematical Imaging and Vision  \textbf{51}(1),
  209--228 (2015)

\bibitem{brachmann2017dsac}
Brachmann, E., Krull, A., Nowozin, S., Shotton, J., Michel, F., Gumhold, S.,
  Rother, C.: Dsac-differentiable ransac for camera localization. In:
  Proceedings of the IEEE Conference on Computer Vision and Pattern Recognition
  (2017)

\bibitem{brachmann2016uncertainty}
Brachmann, E., Michel, F., Krull, A., Ying~Yang, M., Gumhold, S., et~al.:
  Uncertainty-driven 6d pose estimation of objects and scenes from a single rgb
  image. In: Proceedings of the IEEE Conference on Computer Vision and Pattern
  Recognition. pp. 3364--3372 (2016)

\bibitem{brachmann2018learning}
Brachmann, E., Rother, C.: Learning less is more-6d camera localization via 3d
  surface regression. In: Proceedings of the IEEE Conference on Computer Vision
  and Pattern Recognition. pp. 4654--4662 (2018)

\bibitem{brahmbhatt2018geometry}
Brahmbhatt, S., Gu, J., Kim, K., Hays, J., Kautz, J.: Geometry-aware learning
  of maps for camera localization. In: Proceedings of the IEEE Conference on
  Computer Vision and Pattern Recognition. pp. 2616--2625 (2018)

\bibitem{breiman2001random}
Breiman, L.: Random forests. Machine learning  \textbf{45}(1),  5--32 (2001)

\bibitem{bui2018bmvc}
Bui, M., Albarqouni, S., Ilic, S., Navab, N.: Scene coordinate and
  correspondence learning for image-based localization. In: British Machine
  Vision Conference (BMVC) (2018)

\bibitem{busam2017camera}
Busam, B., Birdal, T., Navab, N.: Camera pose filtering with local regression
  geodesics on the riemannian manifold of dual quaternions. In: Proceedings of
  the IEEE International Conference on Computer Vision Workshops. pp.
  2436--2445 (2017)

\bibitem{cadena2016past}
Cadena, C., Carlone, L., Carrillo, H., Latif, Y., Scaramuzza, D., Neira, J.,
  Reid, I., Leonard, J.J.: Past, present, and future of simultaneous
  localization and mapping: Toward the robust-perception age. IEEE Transactions
  on robotics  \textbf{32}(6) (2016)

\bibitem{cayley1846quelques}
Cayley, A.: Sur quelques propri{\'e}t{\'e}s des d{\'e}terminants gauches.
  Journal f{\"u}r die reine und angewandte Mathematik  \textbf{32},  119--123
  (1846)

\bibitem{clark2017vidloc}
Clark, R., Wang, S., Markham, A., Trigoni, N., Wen, H.: Vidloc: A deep
  spatio-temporal model for 6-dof video-clip relocalization. In: Proceedings of
  the IEEE Conference on Computer Vision and Pattern Recognition (2017)

\bibitem{corona2018pose}
Corona, E., Kundu, K., Fidler, S.: Pose estimation for objects with rotational
  symmetry. In: 2018 IEEE/RSJ International Conference on Intelligent Robots
  and Systems (IROS). pp. 7215--7222. IEEE (2018)

\bibitem{cui2019multimodal}
Cui, H., Radosavljevic, V., Chou, F.C., Lin, T.H., Nguyen, T., Huang, T.K.,
  Schneider, J., Djuric, N.: Multimodal trajectory predictions for autonomous
  driving using deep convolutional networks. In: 2019 International Conference
  on Robotics and Automation (ICRA). pp. 2090--2096. IEEE (2019)

\bibitem{Deng2019}
Deng, H., Birdal, T., Ilic, S.: 3d local features for direct pairwise
  registration. In: The IEEE Conference on Computer Vision and Pattern
  Recognition (CVPR) (June 2019)

\bibitem{deng2009imagenet}
Deng, J., Dong, W., Socher, R., Li, L.J., Li, K., Fei-Fei, L.: Imagenet: A
  large-scale hierarchical image database. In: 2009 IEEE conference on computer
  vision and pattern recognition. pp. 248--255. Ieee (2009)

\bibitem{durrant2006simultaneous}
Durrant-Whyte, H., Bailey, T.: Simultaneous localization and mapping: part i.
  IEEE robotics \& automation magazine  \textbf{13}(2),  99--110 (2006)

\bibitem{falorsi2019reparameterizing}
Falorsi, L., de~Haan, P., Davidson, T.R., Forr{\'e}, P.: Reparameterizing
  distributions on lie groups. arXiv preprint arXiv:1903.02958  (2019)

\bibitem{feng20166d}
Feng, W., Tian, F.P., Zhang, Q., Sun, J.: 6d dynamic camera relocalization from
  single reference image. In: Proceedings of the IEEE Conference on Computer
  Vision and Pattern Recognition. pp. 4049--4057 (2016)

\bibitem{firman2018diversenet}
Firman, M., Campbell, N.D., Agapito, L., Brostow, G.J.: Diversenet: When one
  right answer is not enough. In: Proceedings of the IEEE Conference on
  Computer Vision and Pattern Recognition. pp. 5598--5607 (2018)

\bibitem{fischler1981random}
Fischler, M.A., Bolles, R.C.: Random sample consensus: a paradigm for model
  fitting with applications to image analysis and automated cartography.
  Communications of the ACM  \textbf{24}(6),  381--395 (1981)

\bibitem{gal2016dropout}
Gal, Y., Ghahramani, Z.: Dropout as a bayesian approximation: Representing
  model uncertainty in deep learning. In: international conference on machine
  learning. pp. 1050--1059 (2016)

\bibitem{Gilitschenski2020Deep}
Gilitschenski, I., Sahoo, R., Schwarting, W., Amini, A., Karaman, S., Rus, D.:
  Deep orientation uncertainty learning based on a bingham loss. In:
  International Conference on Learning Representations (2020)

\bibitem{glover2014}
Glover, J., Kaelbling, L.P.: Tracking the spin on a ping pong ball with the
  quaternion bingham filter. In: 2014 IEEE International Conference on Robotics
  and Automation (ICRA). pp. 4133--4140 (May 2014)

\bibitem{glover2012monte}
Glover, J., Bradski, G., Rusu, R.B.: Monte carlo pose estimation with
  quaternion kernels and the bingham distribution. In: Robotics: science and
  systems (2012)

\bibitem{glover2014quaternion}
Glover, J.M.: The quaternion Bingham distribution, 3D object detection, and
  dynamic manipulation. Ph.D. thesis, Massachusetts Institute of Technology
  (2014)

\bibitem{grassia1998}
Grassia, F.S.: Practical parameterization of rotations using the exponential
  map. Journal of graphics tools  \textbf{3}(3),  29--48 (1998)

\bibitem{guo2017calibration}
Guo, C., Pleiss, G., Sun, Y., Weinberger, K.Q.: On calibration of modern neural
  networks. In: Proceedings of the 34th International Conference on Machine
  Learning-Volume 70. pp. 1321--1330. JMLR. org (2017)

\bibitem{guzman2012multiple}
Guzman-Rivera, A., Batra, D., Kohli, P.: Multiple choice learning: Learning to
  produce multiple structured outputs. In: Advances in Neural Information
  Processing Systems. pp. 1799--1807 (2012)

\bibitem{haarbach2018survey}
Haarbach, A., Birdal, T., Ilic, S.: Survey of higher order rigid body motion
  interpolation methods for keyframe animation and continuous-time trajectory
  estimation. In: 3D Vision (3DV), 2018 Sixth International Conference on. pp.
  381--389. IEEE (2018). \doi{10.1109/3DV.2018.00051}

\bibitem{he2016deep}
He, K., Zhang, X., Ren, S., Sun, J.: Deep residual learning for image
  recognition. In: Proceedings of the IEEE conference on computer vision and
  pattern recognition. pp. 770--778 (2016)

\bibitem{carl1995}
Herz, C.S.: Bessel functions of matrix argument. Annals of Mathematics
  \textbf{61}(3),  474--523 (1955), \url{http://www.jstor.org/stable/1969810}

\bibitem{hinterstoisser2012model}
Hinterstoisser, S., Lepetit, V., Ilic, S., Holzer, S., Bradski, G., Konolige,
  K., Navab, N.: Model based training, detection and pose estimation of
  texture-less 3d objects in heavily cluttered scenes. In: Asian conference on
  computer vision. Springer (2012)

\bibitem{horaud1989analytic}
Horaud, R., Conio, B., Leboulleux, O., Lacolle, B.: An analytic solution for
  the perspective 4-point problem. In: Proceedings CVPR'89: IEEE Computer
  Society Conference on Computer Vision and Pattern Recognition. IEEE (1989)

\bibitem{kendall2016modelling}
Kendall, A., Cipolla, R.: Modelling uncertainty in deep learning for camera
  relocalization. In: 2016 IEEE international conference on Robotics and
  Automation (ICRA). pp. 4762--4769. IEEE (2016)

\bibitem{kendall2017geometric}
Kendall, A., Cipolla, R., et~al.: Geometric loss functions for camera pose
  regression with deep learning. In: Proc. CVPR. vol.~3, p.~8 (2017)

\bibitem{kendall2017uncertainties}
Kendall, A., Gal, Y.: What uncertainties do we need in bayesian deep learning
  for computer vision? In: Advances in neural information processing systems
  (2017)

\bibitem{kendall2015posenet}
Kendall, A., Grimes, M., Cipolla, R.: Posenet: A convolutional network for
  real-time 6-dof camera relocalization. In: Proceedings of the IEEE
  international conference on computer vision. pp. 2938--2946 (2015)

\bibitem{kingma2014adam}
Kingma, D.P., Ba, J.: Adam: A method for stochastic optimization. arXiv
  preprint arXiv:1412.6980  (2014)

\bibitem{kingma2013auto}
Kingma, D.P., Welling, M.: Auto-encoding variational bayes. arXiv preprint
  arXiv:1312.6114  (2013)

\bibitem{kume2005saddlepoint}
Kume, A., Wood, A.T.: Saddlepoint approximations for the bingham and
  fisher--bingham normalising constants. Biometrika  \textbf{92}(2),  465--476
  (2005)

\bibitem{kurz2013recursive}
Kurz, G., Gilitschenski, I., Julier, S., Hanebeck, U.D.: Recursive estimation
  of orientation based on the bingham distribution. In: Information Fusion
  (FUSION), 2013 16th International Conference on. pp. 1487--1494. IEEE (2013)

\bibitem{kurz2017directional}
Kurz, G., Gilitschenski, I., Pfaff, F., Drude, L., Hanebeck, U.D., Haeb-Umbach,
  R., Siegwart, R.Y.: Directional statistics and filtering using
  libdirectional. arXiv preprint arXiv:1712.09718  (2017)

\bibitem{labbe2019rtab}
Labb{\'e}, M., Michaud, F.: Rtab-map as an open-source lidar and visual
  simultaneous localization and mapping library for large-scale and long-term
  online operation. Journal of Field Robotics  \textbf{36}(2),  416--446 (2019)

\bibitem{makansi2019overcoming}
Makansi, O., Ilg, E., Cicek, O., Brox, T.: Overcoming limitations of mixture
  density networks: A sampling and fitting framework for multimodal future
  prediction. In: Proceedings of the IEEE Conference on Computer Vision and
  Pattern Recognition. pp. 7144--7153 (2019)

\bibitem{manhardt2019explaining}
Manhardt, F., Arroyo, D.M., Rupprecht, C., Busam, B., Birdal, T., Navab, N.,
  Tombari, F.: Explaining the ambiguity of object detection and 6d pose from
  visual data. In: International Conference of Computer Vision. IEEE/CVF (2019)

\bibitem{eitan2016}
Marder-Eppstein, E.: Project tango. pp. 25--25 (07 2016)

\bibitem{mardia2009directional}
Mardia, K.V., Jupp, P.E.: Directional statistics. John Wiley \& Sons (2009)

\bibitem{massiceti2017random}
Massiceti, D., Krull, A., Brachmann, E., Rother, C., Torr, P.H.: Random forests
  versus neural networks—what's best for camera localization? In: 2017 IEEE
  International Conference on Robotics and Automation (ICRA). IEEE (2017)

\bibitem{mclachlan1988mixture}
McLachlan, G.J., Basford, K.E.: Mixture models: Inference and applications to
  clustering, vol.~84. M. Dekker New York (1988)

\bibitem{morawiec1996rodrigues}
Morawiec, A., Field, D.: Rodrigues parameterization for orientation and
  misorientation distributions. Philosophical Magazine A  \textbf{73}(4),
  1113--1130 (1996)

\bibitem{murray1994}
Murray, R.M.: A mathematical introduction to robotic manipulation. CRC press
  (1994)

\bibitem{pytorch}
Paszke, A., Gross, S., Chintala, S., Chanan, G., Yang, E., DeVito, Z., Lin, Z.,
  Desmaison, A., Antiga, L., Lerer, A.: Automatic differentiation in {PyTorch}.
  In: NIPS Autodiff Workshop (2017)

\bibitem{peretroukhin2019probabilistic}
Peretroukhin, V., Wagstaff, B., Giamou, M., Kelly, J.: Probabilistic regression
  of rotations using quaternion averaging and a deep multi-headed network.
  arXiv preprint arXiv:1904.03182  (2019)

\bibitem{piasco2018survey}
Piasco, N., Sidib{\'e}, D., Demonceaux, C., Gouet-Brunet, V.: A survey on
  visual-based localization: On the benefit of heterogeneous data. Pattern
  Recognition  \textbf{74},  90--109 (2018)

\bibitem{pitteri2019object}
Pitteri, G., Ramamonjisoa, M., Ilic, S., Lepetit, V.: On object symmetries and
  6d pose estimation from images. In: 3D Vision (3DV). IEEE (2019)

\bibitem{prokudin2018deep}
Prokudin, S., Gehler, P., Nowozin, S.: Deep directional statistics: Pose
  estimation with uncertainty quantification. In: Proceedings of the European
  Conference on Computer Vision (ECCV). pp. 534--551 (2018)

\bibitem{qi2019deep}
Qi, C.R., Litany, O., He, K., Guibas, L.J.: Deep hough voting for 3d object
  detection in point clouds. In: The IEEE International Conference on Computer
  Vision (ICCV) (October 2019)

\bibitem{riedel2016multi}
Riedel, S., Marton, Z.C., Kriegel, S.: Multi-view orientation estimation using
  bingham mixture models. In: 2016 IEEE international conference on automation,
  quality and testing, robotics (AQTR). pp.~1--6. IEEE (2016)

\bibitem{rubner2000earth}
Rubner, Y., Tomasi, C., Guibas, L.J.: The earth mover's distance as a metric
  for image retrieval. International journal of computer vision
  \textbf{40}(2),  99--121 (2000)

\bibitem{rupprecht2017learning}
Rupprecht, C., Laina, I., DiPietro, R., Baust, M., Tombari, F., Navab, N.,
  Hager, G.D.: Learning in an uncertain world: Representing ambiguity through
  multiple hypotheses. In: Proceedings of the IEEE International Conference on
  Computer Vision. pp. 3591--3600 (2017)

\bibitem{ILSVRC15}
Russakovsky, O., Deng, J., Su, H., Krause, J., Satheesh, S., Ma, S., Huang, Z.,
  Karpathy, A., Khosla, A., Bernstein, M., Berg, A.C., Fei-Fei, L.: {ImageNet
  Large Scale Visual Recognition Challenge}. International Journal of Computer
  Vision (IJCV)  \textbf{115}(3),  211--252 (2015).
  \doi{10.1007/s11263-015-0816-y}

\bibitem{salas2013slam++}
Salas-Moreno, R.F., Newcombe, R.A., Strasdat, H., Kelly, P.H., Davison, A.J.:
  Slam++: Simultaneous localisation and mapping at the level of objects. In:
  Proceedings of the IEEE conference on computer vision and pattern
  recognition. pp. 1352--1359 (2013)

\bibitem{salimans2016improved}
Salimans, T., Goodfellow, I., Zaremba, W., Cheung, V., Radford, A., Chen, X.:
  Improved techniques for training gans. In: Advances in neural information
  processing systems. pp. 2234--2242 (2016)

\bibitem{sattler2015hyperpoints}
Sattler, T., Havlena, M., Radenovic, F., Schindler, K., Pollefeys, M.:
  Hyperpoints and fine vocabularies for large-scale location recognition. In:
  Proceedings of the IEEE International Conference on Computer Vision. pp.
  2102--2110 (2015)

\bibitem{sattler2019understanding}
Sattler, T., Zhou, Q., Pollefeys, M., Leal-Taixe, L.: Understanding the
  limitations of cnn-based absolute camera pose regression. In: Proceedings of
  the IEEE Conference on Computer Vision and Pattern Recognition. pp.
  3302--3312 (2019)

\bibitem{schonberger2016structure}
Schonberger, J.L., Frahm, J.M.: Structure-from-motion revisited. In:
  Proceedings of the IEEE Conference on Computer Vision and Pattern Recognition
  (2016)

\bibitem{shotton2013scene}
Shotton, J., Glocker, B., Zach, C., Izadi, S., Criminisi, A., Fitzgibbon, A.:
  Scene coordinate regression forests for camera relocalization in rgb-d
  images. In: Proceedings of the IEEE Conference on Computer Vision and Pattern
  Recognition. pp. 2930--2937 (2013)

\bibitem{Simonyan14c}
Simonyan, K., Zisserman, A.: Very deep convolutional networks for large-scale
  image recognition. CoRR  \textbf{abs/1409.1556} (2014)

\bibitem{subbarao2009nonlinear}
Subbarao, R., Meer, P.: Nonlinear mean shift over riemannian manifolds.
  International journal of computer vision  \textbf{84}(1), ~1 (2009)

\bibitem{ley2018directional}
Suvrit, S., Ley, C., Verdebout, T.: Directional statistics in machine learning:
  A brief review. In: Applied Directional Statistics. Chapman and Hall/CRC
  (2018)

\bibitem{szegedy2016rethinking}
Szegedy, C., Vanhoucke, V., Ioffe, S., Shlens, J., Wojna, Z.: Rethinking the
  inception architecture for computer vision. In: Proceedings of the IEEE
  conference on computer vision and pattern recognition. pp. 2818--2826 (2016)

\bibitem{ullman1979interpretation}
Ullman, S.: The interpretation of structure from motion. Proceedings of the
  Royal Society of London. Series B. Biological Sciences  \textbf{203}(1153),
  405--426 (1979)

\bibitem{valentin2015exploiting}
Valentin, J., Nie{\ss}ner, M., Shotton, J., Fitzgibbon, A., Izadi, S., Torr,
  P.H.: Exploiting uncertainty in regression forests for accurate camera
  relocalization. In: Proceedings of the IEEE Conference on Computer Vision and
  Pattern Recognition. pp. 4400--4408 (2015)

\bibitem{yamaji2016genetic}
Yamaji, A.: Genetic algorithm for fitting a mixed bingham distribution to 3d
  orientations: a tool for the statistical and paleostress analyses of fracture
  orientations. Island Arc  \textbf{25}(1),  72--83 (2016)

\bibitem{zakharov2019dpod}
Zakharov, S., Shugurov, I., Ilic, S.: Dpod: 6d pose object detector and
  refiner. In: The IEEE International Conference on Computer Vision (ICCV)
  (2019)

\bibitem{zeisl2015camera}
Zeisl, B., Sattler, T., Pollefeys, M.: Camera pose voting for large-scale
  image-based localization. In: Proceedings of the IEEE International
  Conference on Computer Vision. pp. 2704--2712 (2015)

\bibitem{Zhou2018}
Zhou, Q.Y., Park, J., Koltun, V.: {Open3D}: {A} modern library for {3D} data
  processing. arXiv:1801.09847  (2018)

\bibitem{zhou2019continuity}
Zhou, Y., Barnes, C., Lu, J., Yang, J., Li, H.: On the continuity of rotation
  representations in neural networks. In: Proceedings of the IEEE Conference on
  Computer Vision and Pattern Recognition. pp. 5745--5753 (2019)

\bibitem{zolfaghari2019learning}
Zolfaghari, M., {\c{C}}i{\c{c}}ek, {\"O}., Ali, S.M., Mahdisoltani, F., Zhang,
  C., Brox, T.: Learning representations for predicting future activities.
  arXiv:1905.03578  (2019)

\end{thebibliography}
